\PassOptionsToPackage{pdfpagelabels=false}{hyperref} 

\documentclass{edm_article}

\usepackage{adjustbox}
\usepackage{tabularx,booktabs}

\usepackage{enumitem}
\usepackage{arydshln}

\newcommand{\KCluster}{\texttt{KCluster}}

\usepackage{hyperref}
\begin{document}

\title{KCluster: An LLM-based Clustering Approach to Knowledge Component Discovery}

% Submissions for EDM are double-blind: please do not include any author names or affiliations in the submission. 
% Anonymous authors:
% \numberofauthors{1}
% \author{
% Anonymous\\
%        \affaddr{Anonymous Institution}\\
%        \email{anonymous@anonymous.edu}
% }
%An example of how to include
% multiple authors is below for after the paper has been accepted.


\numberofauthors{3} %  in this sample file, there are a *total*
% % of EIGHT authors. SIX appear on the 'first-page' (for formatting
% % reasons) and the remaining two appear in the \additionalauthors section.
% %
\author{
% 1st. author
\alignauthor
Yumou Wei\\
       \affaddr{Human-Computer Interaction Institute}\\
       \affaddr{Carnegie Mellon University}\\
       % \affaddr{Pittsburgh, PA}\\
       \email{yumouw@cs.cmu.edu}
% 2nd. author
\alignauthor
Paulo Carvalho\\
       \affaddr{Human-Computer Interaction Institute}\\
       \affaddr{Carnegie Mellon University}\\
       % \affaddr{Pittsburgh, PA}\\
       \email{pcarvalh@cs.cmu.edu}
% 3rd. author
\alignauthor 
John Stamper\\
       \affaddr{Human-Computer  Interaction Institute}\\
       \affaddr{Carnegie Mellon University}\\
       % \affaddr{Pittsburgh, PA}\\
       \email{jstamper@cs.cmu.edu}
}
% \and  % use '\and' if you need 'another row' of author names
% % 4th. author
% \alignauthor Lawrence P. Leipuner\\
%        \affaddr{Brookhaven Laboratories}\\
%        \affaddr{Brookhaven National Lab}\\
%        \affaddr{P.O. Box 5000}\\
%        \email{lleipuner@researchlabs.org}
% % 5th. author
% \alignauthor Sean Fogarty\\
%        \affaddr{NASA Ames Research Center}\\
%        \affaddr{Moffett Field}\\
%        \affaddr{California 94035}\\
%        \email{fogartys@amesres.org}
% % 6th. author
% \alignauthor Charles Palmer\\
%        \affaddr{Palmer Research Laboratories}\\
%        \affaddr{8600 Datapoint Drive}\\
%        \affaddr{San Antonio, Texas 78229}\\
%        \email{cpalmer@prl.com}
% }
% % There's nothing stopping you putting the seventh, eighth, etc.
% % author on the opening page (as the 'third row') but we ask,
% % for aesthetic reasons that you place these 'additional authors'
% % in the \additional authors block, viz.
% \additionalauthors{Additional authors: John Smith (The Th{\o}rv{\"a}ld Group,
% email: {\texttt{jsmith@affiliation.org}}) and Julius P.~Kumquat
% (The Kumquat Consortium, email: {\texttt{jpkumquat@consortium.net}}).}
% \date{30 July 1999}
% Just remember to make sure that the TOTAL number of authors
% is the number that will appear on the first page PLUS the
% number that will appear in the \additionalauthors section.

\maketitle
    
\begin{abstract}
Educators evaluate student knowledge using knowledge component (KC) models that map assessment questions to KCs. Still, designing KC models for large question banks remains an insurmountable challenge for instructors who need to analyze each question by hand. The growing use of Generative AI in education is expected only to aggravate this chronic deficiency of expert-designed KC models, as course engineers designing KCs struggle to keep up with the pace at which questions are generated. In this work, we propose KCluster, a novel KC discovery algorithm based on identifying clusters of congruent questions according to a new similarity metric induced by a large language model (LLM). We demonstrate in three datasets that an LLM can create an effective metric of question similarity, which a clustering algorithm can use to create KC models from questions with minimal human effort. Combining the strengths of LLM and clustering, KCluster generates descriptive KC labels and discovers KC models that predict student performance better than the best expert-designed models available. In anticipation of future work, we illustrate how KCluster can reveal insights into difficult KCs and suggest improvements to instruction. 
\end{abstract}

 % Replace with your own 3-5 keywords

\section{Introduction}

Real knowledge is to know the extent of one's ignorance---as Confucius reflected on his epistemology. One way edu\-cators can evaluate student knowledge, according to the Knowledge-Learning-Instruction (KLI) framework~\cite{kli}, is by developing cognitive models that map assessment items (or questions) to knowledge components. A \textbf{knowledge component} (KC) is a unit of cognitive function or structure that a student acquires through learning~\cite{kli}, representing specific information, concepts, or skills that a student needs to solve a task or a problem---a student must know how to ``use guide words'' before determining whether ``guess'' can be found on a dictionary page marked with ``garage'' and ``goose''. With a well-designed cognitive model (or \textbf{KC model}), instructors can divide a complex topic into simpler and more manageable milestones that help track student learning~\cite{corbett_knowledge_1994}, identify learning sub-goals with which students struggle~\cite{stamper_kc}, and organize instruction events to promote knowledge transfer~\cite{koedinger_testing_2016}.

Despite the numerous benefits of a well-designed KC model, mapping assessment questions to KCs still remains an insurmountable challenge for instructors and instructional designers who are overwhelmed by the sheer amount of questions that each need to be analyzed by hand. Cognitive Task Analysis (CTA)~\cite{Clark2008}, the \emph{de facto} best manual approach to KC discovery, incurs considerable labor and time costs that prevent schools and teachers from gaining equitable access; therefore, many datasets that naturally occur from students interacting with educational technologies lack \textbf{expert-designed KC models}. For example, nearly 60\% of the 4,639 datasets available in DataShop~\cite{datashop}---the largest educational data repository---do not contain more significant KC models than the default \texttt{Single-KC} and \texttt{Unique-step} models that are only intended to serve as benchmarks\footnote{Based on DataShop administrators' response to our inquiry in December 2024}. This absence of expert-designed KCs limits the analytics that can be conducted and the educational insights that such data can provide. Furthermore, we expect that the increasing adoption of Generative AI (GenAI) in education can only exacerbate this deficiency, as learning engineers developing KCs struggle to keep up with the pace at which questions are produced by GenAI and become even less likely to provide quality KCs. 

This chronic deficiency of expert-designed KCs in large question banks, aggravated by the accelerating use of GenAI in education, calls for a new effective KC discovery algorithm that can automatically extract KCs from abundant question content with minimal burden on instructors. A notable approach, SMART~\cite{Matsuda_Wood_Shrivastava_Shimmei_Bier_2022}, extracts KCs from instructional content based on the assumption that a cluster of linguistically similar texts shares the same KC. SMART applies $k$-means \textbf{clustering} to the TF-IDF embeddings of instructional texts and obtains descriptive KC labels using a keyword extraction algorithm called TextRank~\cite{mihalcea-tarau-2004-textrank}. Although shown in two science datasets to create KC models that predict student responses better than expert-designed models do, SMART still requires a course engineer to specify the number of KCs to discover---a hyperparameter that the authors reported has a statistically significant impact on how well SMART fits to the student data; moreover, identifying each KC with short keywords, SMART tends to produce coarse labels that result in \emph{identical} labels for what experts believe should be \emph{separate} KCs. A more recent approach~\cite{moore_kc} uses a \textbf{large language model} (LLM) to identify KCs for multiple-choice questions. The authors implemented two strategies---simulated expert and simulated textbook---that encourage the LLM to generate descriptive KC labels based on question content. Although in an evaluation study involving three participants, the majority preferred the LLM-generated KC labels to those crafted by experts for more than 60\% of the evaluated questions, this LLM-based approach, contrary to SMART, produces \emph{slightly different} labels for questions that experts believe should belong to the \emph{same} KC, as acknowledged by the authors. The two current divergent approaches to KC discovery beg the question: \textbf{Will a hybrid of clustering and LLM produce synergy in extracting KCs from questions?}

In this work, we propose~\KCluster\footnote{Pronounced the same as ``cluster'', \texttt{KCluster} is freely available at \url{https://github.com/weiyumou/KCluster}.}, an unsupervised \textbf{KC} discovery algorithm based on identifying \textbf{cluster}s of \emph{congruent} questions according to a novel similarity metric induced by an LLM. By extending word collocations to questions, we developed a novel concept called \textbf{question congruity} that quantifies the similarity of two questions by the likelihood of their co-occurrence, and devised an algorithm that uses LLM as a probability machine to compute the required text probabilities \textbf{without retraining or finetuning the LLM}. Combining the strengths of LLM and clustering, \texttt{KCluster} uses \emph{Phi-2}~\cite{hughes_phi-2_2023} (an LLM) to measure question congruity and generate descriptive KC labels, and uses \emph{affinity propagation}~\cite{doi:10.1080/01621459.1983.10478008} (a clustering algorithm) to identify clusters of congruent questions, each corresponding to a KC. We validated \texttt{KCluster} on three datasets related to science and e-learning, two of which contain student response data, giving affirmative answers to our three research questions (RQs):
\begin{itemize}[leftmargin=*]
   \item RQ-1: Does \texttt{KCluster} align with expert-designed KC mo\-dels? (Section~\ref{sec: alignment})
    \item RQ-2: Does \texttt{KCluster} enable accurate prediction of student responses? (Section~\ref{sec: afm})
    \item RQ-3: Does \texttt{KCluster} reveal insights about problematic KCs? (Section~\ref{sec: insights})
\end{itemize}

Through our comprehensive evaluation comparing \texttt{KCluster} to three other competitive methods on large question banks and student data, we demonstrate that \textbf{an LLM can create a new, effective measure of similarity between two arbitrary questions, which a clustering algorithm can use to extract KCs from questions automatically, without elaborate retraining, finetuning, or prompt engineering}. The main contributions of our research include: 1) a novel measure of question similarity, 2) an algorithm to compute the new similarity metric using LLM, and 3) an effective approach to extract descriptive KC labels from question content.   
% Every successful piece of non-fiction writing should leave the readers with ONE idea that they have not thought about before.

\section{Literature Review}
A comprehensive review of the literature on KC discovery is necessary to show how \texttt{KCluster} connects to and builds on current approaches. We classify the approaches into three categories based on the amount of manual work required and review them in decreasing order of human involvement. 
% Approaches differ from each other along three important dimensions: whether an approach obviates human input, produces descriptive KC labels, and is validated on student data.
% Table~\ref{tab: related-work} offers a side-by-side comparison of all the methods along three dimensions: whether a method obviates human input, produces descriptive KC labels, and is validated on student data.

\subsection{Manual Approaches}
Manual approaches rely solely on the expertise of an instructional designer to identify KCs. Although a teacher could review and label each question with a KC, a more systematic approach is through Cognitive Task Analysis (CTA)~\cite{Clark2008}, where instructional experts are asked to elucidate their mental processes in solving problems during a think-aloud interview. A notable CTA approach is Difficulty Factors Assess\-ment (DFA)~\cite{HeffernanKoedinger1998,Koedinger2004}, based on the assumption that students should perform similarly on questions concerning the same KC---therefore, any performance discrepancy is due to a hidden KC yet to be discovered. For example, using DFA, researchers identified a new KC (about comprehending the symbolic representation of quantitative relations) that explained why beginning algebra students performed worse on algebra problems presented with mathematical symbols than on problems embedded in a hypothetical story, illuminating the effect of problem presentation on learning that had been overlooked~\cite{Koedinger2004}. Although CTA is known to improve instru\-ction~\cite{Koedinger2010-KOESLL}, the outcome is highly sensitive to the CTA methods used and the instructional context considered~\cite{doi:10.1177/1555343412474821}. Moreover, CTA relies heavily on experts to make subjective decisions and therefore incurs considerable labor and time costs that prevent CTA from scaling to large question banks readily available with GenAI. (Semi-)automated approaches, however, alleviate the scalability problem by minimizing human involvement and learning KC models from data.
% [crowdsourcing] 

\subsection{Semi-automated Approaches}
Semi-automated approaches refine an expert-designed KC model with data-driven methods. A notable approach~\cite{stamper_kc} extends DFA with a statistical model of student data to identify problematic KCs worth improving; by analyzing a difficult KC identified from data, researchers uncovered three hidden KCs for geometry area learning
% refined a KC model for geometry area learning by splitting a composite KC into three finer ones, 
and obtained a better prediction of student performance. In a sequel~\cite{10.1007/978-3-642-39112-5_43}, researchers reaffirmed the efficacy of this data-driven DFA approach by redesigning a cognitive tutor for teaching geo\-metry and showing improvements in student learning. An alternate approach, Learning Factors Analysis (LFA)~\cite{lfa}, further automates DFA by using the A\textsuperscript{$\ast$} algorithm~\cite{RussellNorvig2020} to search for better KC models based on a list of difficult factors that experts think are absent from the current model. In an evaluation study~\cite{koedinger_automated_2012} researchers found LFA improve KC models across ten datasets of various domains and closed the development-test-redesign loop in a sequel~\cite{Liu_Koedinger_2017} that redesigned a tutoring system using LFA-generated insights. Although semi-automated approaches are grounded on student data, they rely on expert-designed KC models to produce descriptive KC labels, calling for more automated approaches that eliminate human input.   

\subsection{Automated Approaches}
Automated approaches develop new KC models from scratch and do not require human input beyond a few hyperparameters. The Q-matrix method~\cite{Barnes2005TheQM} and its sequels using matrix factorization~\cite{10.1007/978-3-642-30950-2_58,10.1007/978-3-642-39112-5_45,JMLR:v15:lan14a} search for a KC model that best predicts student responses to questions. A closely related class of approaches discovers KCs as part of a statistical model learned from data---one method~\cite{liu_data-driven_2012} creates KC models through a DINA model~\cite{dina}, while dAFM~\cite{pardos_dafm_2018} and SparFAE~\cite{2022.EDM-long-papers.2}, both using neural networks, estimate Q-matrices via an AFM~\cite{afm} and an IRT model~\cite{HambletonSwaminathan1985}; other similar approaches have explored Hidden Markov Model~\cite{GonzlezBrenes2013WhatAW} and extended to identifying KCs in programming problems~\cite{shi_kc-finder_2023}. These automated approaches based on statistical learning, although capable of identifying KCs without human intervention, still require reference to an expert-designed KC model to produce descriptive KC labels (otherwise, they produce nominal labels such as ``KC-15'', which provides no instructional insights); therefore, they are better suited for unsupervised \emph{KC refinement} than automatic KC discovery.

A unique class of automated approaches that can produce descriptive KC labels without a reference model extracts KCs from \emph{instructional content} such as textbooks. For example, SimStudent-based approach~\cite{DBLP:conf/edm/LiCKM11,li_general_2013} iteratively associates predefined skill labels with \emph{problem-solving demonstrations} and creates new KC labels if necessary; similarly, researchers have explored a term-matching approach to extract concepts from \emph{student explanations} for math problems~\cite{shabana_unsupervised_2023}. Another approach, FACE~\cite{chau_automatic_2021}, identifies concepts from \emph{adaptive textbooks} based on an extensive list of hand-engineered features. All these approaches, however, require a list of key skills or concepts specified by experts beforehand. A notable approach that does not require human input, SMART~\cite{Matsuda_Wood_Shrivastava_Shimmei_Bier_2022}, extracts KCs from instructional texts and questions by clustering similar texts encoded as TF-IDF vectors. The $k$-means clustering algorithm was applied to both the embedding vectors and their cosine similarity, although no significant differences were observed; the researchers then applied TextRank~\cite{mihalcea-tarau-2004-textrank} to extract keywords from each of the $k$ clusters to use as KCs. Although SMART was validated on two science datasets to create quality KC models, it still required an expert to specify $k$, the number of KCs to discover, and the keywords identified by TextRank were so coarse that resulted in duplicate labels for what experts believe should be distinct KCs. A more recent approach~\cite{moore_kc} uses an LLM to identify KCs from multiple-choice questions by asking the LLM to simulate instructional experts or textbook authors. Although in a three-subject evaluation study, the majority of the evaluators showed preference for LLM-generated KC labels in more than 60\% of the evaluated questions, this approach produces an excessive number of KCs because, in contrast to TextRank, LLM is so capable that it generates slightly different labels for questions that experts believe should belong to the same KC. The two divergent approaches suggest that clustering, capable of uncovering latent question structures, and LLM, capable of generating descriptive KC labels, can form synergy in KC discovery.

\section{Methods}

We propose, evaluate, and compare three classes of automated KC extraction methods, each of which extends the preceding method and builds upon a large language model (LLM). The LLM we used in this work is Phi-2~\cite{hughes_phi-2_2023} from Microsoft, a lightweight open-source model trained on high-quality textbook-like data~\cite{gunasekar2023textbooks} and potentially suited for educational data mining. We used the Phi-2 distribution freely available through HuggingFace~\cite{wolf-etal-2020-transformers} and used PyTorch~\cite{10.5555/3454287.3455008} for our custom implementation. Phi-2 was deployed to a computing cluster with access to NVIDIA A40 GPUs.

Although having fewer parameters (2.7B) than most mainstream models, Phi-2 is ideal for building our KC discovery algorithms and providing resource-constrained institutions with equitable access to GenAI tools because it offers a good balance between performance and affordability~\cite{hughes_phi-2_2023}. Our choice of Phi-2 may seem unconventional, when ``GPT'' has nearly become synonymous with LLMs. However, we believe that Phi-2 offers two distinct advantages that make it a compelling choice for our research. First, Phi-2 is an open-source model, which allows us to access its hidden states and output log-probabilities essential for developing our KC extraction methods; as shown in Section~\ref{sec: kcluster}, \texttt{KCluster} requires us to evaluate the log-probability of \emph{any} token, whereas the Open\-AI API\footnote{\url{https://platform.openai.com/docs/api-reference/chat/create\#chat-create-top_logprobs}} only supports the top 20 most likely tokens \emph{that it returns}. Second, under modest hardware requirements, Phi-2 is overall the best LLM with <10B parameters, outperforming Mistral 7B~\cite{jiang2023mistral} and Llama-2 13B~\cite{touvron2023llama2} in math~\cite{cobbe2021gsm8k} and coding~\cite{chen2021evaluating} tasks; smaller or earlier models like BERT~\cite{devlin2019bert} would not have benefited from the extensive pre-training on large textbook-like corpora that made Phi-2 potentially suitable for educational tasks. Building our three KC extraction methods with Phi-2 represents a leading effort to explore the potential of alternative LLMs for educational applications, such as KC discovery. 

% Moreover, through extensive experiments, we discovered Phi-2's distinctive ability to respond to prompts that contain educational content (such as questions), which reinforced our predilection for it.  

\subsection{Concept Extraction}\label{sec: baseline}

\begin{table}[t]
    \centering
    % \captionsetup{labelfont=bf,labelformat=empty,labelname=Listing,labelsep=none} % Customize caption

    \caption{The prompt template used in \texttt{Concept} (left) and a concrete prompt filled with an actual question (right).} 
    \label{tab: baseline_prompt}
\begin{adjustbox}{width=\linewidth}
    \begin{tabular}
{p{0.49\linewidth}p{0.49\linewidth}}
    \toprule
    \vspace{-1.5em}
\begin{verbatim}
Exercise 1:
{question type}:
{stem}
{
    choices
}
Answer: {answer}

Remark:
The above exercise is a 
{question type} question 
that tests whether the 
student understands the 
concept of [extracted KC]
\end{verbatim}   & \vspace{-1.5em} \begin{verbatim}
Exercise 1:
Multiple Choice:
Which is the most flexible?
a) paper
b) ceramic tea cup
c) clay tile
Answer: a) paper

Remark:
The above exercise is a 
multiple-choice question 
that tests whether the 
student understands the 
concept of [extracted KC]
\end{verbatim}     \\  [-1em]
    \bottomrule
    \end{tabular}
\end{adjustbox}
\end{table}

\begin{table*}[ht]
    \centering
    \caption{Code for obtaining question embeddings from Phi-2; speacial markers like \texttt{Exercise 1} are passed to the \texttt{text} parameter of the tokenizer and question texts to \texttt{text\_pair}. Only embeddings of the question texts are obtained.} 
    \label{tab: embed}
% \begin{adjustbox}{width=\textwidth}
    \begin{tabular}
{p{0.97\textwidth}}
    \toprule
    \vspace{-1.5em}
% \begin{lstlisting}[language=Python, frame=none,basicstyle=\ttfamily,showstringspaces=false]
\begin{verbatim}
inputs = tokenizer(text=["Exercise 1:..."], text_pair=["Which is the..."],
                   return_tensors="pt", return_token_type_ids=True, padding=True).to(device)
mask = inputs.pop("token_type_ids").bool() # useful for distinguishing markers from question content
indices = torch.cumsum(mask.sum(-1), dim=-1).tolist() # the starting index of each question's text
last_states = model(**inputs, output_hidden_states=True).hidden_states[-1] # a loaded Phi-2 model
chunks = torch.vsplit(last_states[mask], indices)[:-1]
embeddings = torch.stack([torch.mean(c, dim=0) for c in chunks], dim=0)
\end{verbatim} \\ [-1em]
% \end{lstlisting} \\ [-1em]
    \bottomrule
    \end{tabular}
% \end{adjustbox}
\end{table*}

A straightforward application of LLMs to KC discovery is to extract concepts from questions. In line with previous work using LLMs~\cite{moore_kc}, we explicitly ask Phi-2 to identify the \emph{key concept} that a student must know to answer a question correctly and treat each concept as a KC. Through extensive prompt engineering, we discovered an effective \emph{prompt template}, which allowed us to obtain descriptive and accurate concept labels without elaborate prompting strategies as used in previous work~\cite{moore_kc}. Shown on the left of Table~\ref{tab: baseline_prompt}, the prompt template includes special markers to which Phi-2 is particularly responsive. For example, we discovered that the marker ``\texttt{Exercise 1:}'' followed by \texttt{\{question type\}} prompts Phi-2 to generate a new question in a format that we now adopt in the prompt template (namely, \texttt{stem}, \texttt{choices}, and \texttt{Answer:}). Similarly, ``\texttt{Remark:}'' encourages Phi-2 to write a comment starting with ``The above exercise...'' about the preceding question; therefore, we expanded the remark with more explicit instructions asking Phi-2 to complete generation with the key concept. None of these special markers are officially documented~\cite{hughes_phi-2_2023}, but are discovered from our extensive prompt engineering. On the right of Table~\ref{tab: baseline_prompt} shows a concrete prompt derived from the template by replacing the variables in curly brackets with specific values. We denote this method as \texttt{Concept}.

% Text generation using LLMs often proceeds in multiple steps, and various \emph{decoding strategies} can guide Phi-2 in generating key concepts. 
In generating the key concepts, we adopt a \emph{greedy decoding strategy}, in which Phi-2 always selects the most probable token at each generation step. Moreover, we use beam search to maintain five candidate concepts during generation and apply a length penalty~\cite{wu2016googlesneuralmachinetranslation} to encourage Phi-2 to generate succinct concepts---for the example prompt shown in Table~\ref{tab: baseline_prompt}, Phi-2 produced ``flexibility''. Generation stops when a period or comma appears, and we select the best candidate with the highest probability. As shown in Section~\ref{sec: res_dis}, using concepts as KCs, \texttt{Concept} is a competitive baseline that produces KC labels in reasonable alignment with expert-crafted ones; it is also used by other KC discovery algorithms described hereinafter
to create descriptive KC labels.

\subsection{Semantic Embedding}

\begin{figure*}[t]
    \centering
\includegraphics[width=\linewidth]{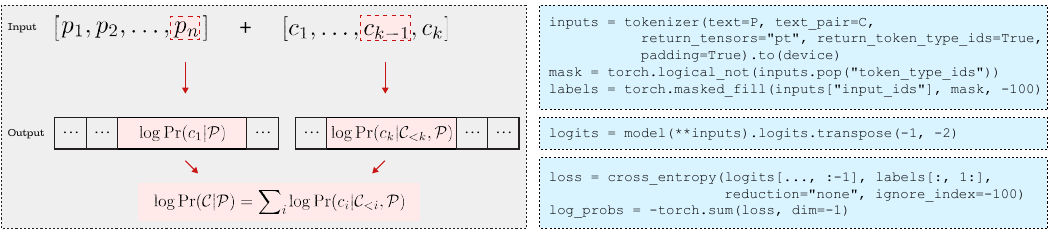}
    \caption{An illustration of the algorithm for computing $\log\Pr(\mathcal{C}| \mathcal{P})$, along with three code snippets for each key step} 
    % the conditional log-probability of a prompt continuation $\mathcal{C}$ given the prompt $\mathcal{P}$}
    \label{fig: next-token}
    \Description{This figure illustrates the algorithm for computing the conditional log-probability of C given P.}
\end{figure*}

A known limitation of \texttt{Concept}, as encountered in previous work~\cite{moore_kc}, is that the LLM can generate \emph{slightly different} KC labels for questions to which an instructional expert would assign the \emph{same} KC---the single and plural forms of the same concept (gas vs. gases), among other trivialities, can result in redundant labels that could have been merged. One approach to reducing such redundancy, as used in SMART~\cite{Matsuda_Wood_Shrivastava_Shimmei_Bier_2022}, is to group similar instructional items by applying a clustering algorithm to their \emph{semantic embeddings} and assign each group to a KC. Depending on which item we convert to embeddings, we introduce two embedding-based methods as enhanced baselines. 
\begin{itemize}[leftmargin=*]
    \item \textbf{Concept embedding}:
A natural extension to \texttt{Concept} is to encode the key concepts extracted by Phi-2 as vectors and assign questions to KCs based on \emph{concept similarity}. Since each concept is a short phrase, we use a state-of-the-art sentence embedding model, Sentence Transformer~\cite{reimers-2019-sentence-bert} with ``\texttt{all-mpnet-base-v2}'' backend that offers the best quality, to produce a vector of 768 dimensions for each concept. We call this method \texttt{Concept-emb}. 

\item \textbf{Question embedding}: An alternative is to encode the questions, which contain more information than the concepts, and group the questions based on \emph{question similarity}. We present questions to Phi-2 using the same prompt template shown on the left of Table~\ref{tab: baseline_prompt} (without \texttt{Remark}), and take the 2560-dimensional average vector of Phi-2's last hidden states before the language-modeling head as question embeddings (a code snippet is listed in Table~\ref{tab: embed} for reference). We call this method \texttt{Question-emb}. 
\end{itemize}

Since the two methods produce embeddings of different sizes, to ensure a fair comparison, we further reduce the embeddings to their similarities. As shown in SMART~\cite{Matsuda_Wood_Shrivastava_Shimmei_Bier_2022}, using similarity rather than embeddings does not affect the quality of the resulting KC models---if not more advantageous. In particular, we use \emph{negative cosine distance}\footnote{It is equivalent to cosine similarity but more compatible with other negative distances that future work may explore.}, defined as $\cos(\mathbf{x}, \mathbf{y}) - 1$ for two vectors $\mathbf{x}$ and $\mathbf{y}$, to quantify the similarity between two embeddings (the values range from -2 to 0, with identical vectors having the largest value of 0).

After we obtain the similarity matrix of the embeddings, we use clustering to identify questions that share similar concepts (as in \texttt{Concept-emb}) or content (as in \texttt{Question-emb}). The clustering algorithm we used is \emph{affinity propagation}~\cite{doi:10.1126/science.1136800}, which does not require the number or the size of the clusters to be pre-specified; instead, it takes as input a matrix describing the affinity between input items and discovers item clusters through optimization. Each cluster is uniquely identified by its central item called ``exemplar'', and the user can specify an initial \emph{preference} for each input item to be an exemplar. 
% , which defaults to the median object affinity if unspecified. 
% To initialize the clusters, , and 
The algorithm is so named because it propagates between items two kinds of \emph{messages} derived from the affinity matrix: at every iteration, an item $i$ sends to another item $j$ a number (the message) reflecting the \emph{responsibility} for $i$ to choose $j$ as an exemplar over others, and receives from $j$ another number indicating $j$'s \emph{availability} to be an exemplar of $i$ with respect to other items that have chosen $j$ as an exemplar. In essence, affinity propagation stimulates the items to compete for being an exemplar and halts when the exemplars (and the clusters) stop changing. In addition to not requiring the number of clusters be specified, affinity propagation accepts affinity measures that are not necessa\-rily a mathematical metric, allowing the use of task-specific measures that are expected to result in better performance. 

We set a uniform preference (using the median affinity of all pairs of input, by default) for each concept or question to be an exemplar. At convergence, affinity propagation produces a nominal cluster label for each input item and a one-to-one mapping of exemplars to clusters. While questions within a cluster are assigned the same KC, for both \texttt{Concept-emb} and \texttt{Question-emb}, we label each question with the concept of its exemplar that we obtained from \texttt{Concept}. In practice, we always run \texttt{Concept} for all questions before running either embedding-based method to ensure that every cluster has a descriptive label, whichever questions become exemplars. If two exemplars have identical concepts, two previously separate KCs may be (unintentionally) merged, but practitioners can always choose whether or not to merge those KCs, depending on whichever leads to better performance. As shown in Section~\ref{sec: res_dis}, using a classic similarity measure (negative cosine distance), \texttt{Question-emb} significantly outperforms \texttt{Concept} and produces less redundant KC labels.

\subsection{KCluster}\label{sec: kcluster}

Using the classic cosine-based metric to measure concept or question similarity misses an opportunity to fully exploit an LLM's capability---after all, producing question embeddings is perhaps not the best use of an LLM. In addition to generating text as in \texttt{Concept}, a large \emph{language model} is also an exceptional ``probability machine'' that can evaluate the probability of an arbitrary piece of text~\cite{jm3}, \textbf{even without retraining or finetuning}. Our main KC extraction method, \texttt{KCluster}, retains the use of affinity propagation to group similar questions, but extends \texttt{Question-emb} with a new measure of question similarity based on text probabilities. We introduce \emph{question congruity}, a new similarity metric derived from quantifying the likelihood of question collocations, and describe
an algorithm that uses Phi-2 to compute the required probabilities.

% An important distinction between~\KCluster~ and existing LLM-based KC extraction methods including our baseline is that~\KCluster~does \textbf{not} just rely on the ``KC'' generated by an LLM; 
% % through a text generation procedure that is difficult to manipulate or interpret. 
% instead, using the underlying architecture of an LLM as a ``probability machine'' that can evaluate the probability of an arbitrary text,~\KCluster~\emph{measures question congruity} and \emph{groups congruent questions} to identify KCs. \KCluster~is so named because it is a \textbf{KC} discovery algorithm based on \textbf{clustering}, which uses our novel concept of question congruity calculated by an LLM to aggregate congruent questions by presumptive KCs. As shown in Section~\ref{sec: alignment}, the additional components included in~\KCluster, for measuring question congruity and clustering questions, allow~\KCluster~to outperform the baseline in aligning with experts and predicting student performance, although both methods share a common procedure to create interpretable KC labels. 

\subsubsection{Collocating questions are congruent}\label{sec: congruity}

In a coherent speech, words are not uttered haphazardly but join other \emph{congruent} words to form \emph{collocations} (e.g., ``data mining''); therefore, if one word makes the other more likely to appear in a sentence than otherwise, the two words are congruent. Since questions are made up of words, the notion of congruity can be extended from words to questions. Based on instructional design principles, we postulate that, as two words collocate in a sentence to form a phrase, \textbf{two questions can co-occur (in a worksheet or an exam paper) if they belong to the same unit, the same lesson, or better still, \emph{the same KC}}. To quantify the collocation of two questions, $q_{s}$ and $q_{t}$, we consider how much \emph{more likely} the presence of $q_{t}$ makes $q_{s}$ to appear, by evaluating the change in log-probabilities of $q_{s}$ with and without $q_{t}$, and defining:
\begin{equation}\label{eqn: pmi}
    \Delta(q_{s}, q_{t}) := \log \Pr(q_{s} | q_{t}) - \log \Pr(q_{s}) % = \log \frac{\Pr(q_{s} | q_{t})}{\Pr(q_{s})}
\end{equation}
The underlying principle is similar to that for words: if one question significantly increases the other question's likelihood of occurrence, in which case $\Delta(q_{s}, q_{t})$ is large positive, the two questions must be highly congruent---they may use a similar language, concern a single topic, or come from the same textbook chapter, all alluding to \textbf{a shared KC}; on the other hand, if $\Delta(q_{s}, q_{t})$ is close to zero or negative, in which case the presence of $q_{t}$ does not improve (or even hurts) the likelihood of $q_{s}$ appearing, the two questions are hardly congruent---they are unlikely to belong to the same KC, since they do not even co-occur often. 
% Thus, by measuring pairwise question congruity as we do with words, we can identify \emph{clusters} of questions that may share the same KC --- the core idea of~\KCluster. 

Equation~\ref{eqn: pmi} only partially quantifies question congruity as it assumes that $q_{t}$ precedes $q_{s}$; however, two questions can also co-occur (and be congruent) when, conversely, $q_{s}$ precedes $q_{t}$. Therefore, we take a step further to \textbf{define} question congruity formally as a symmetric quantity that equally weighs both cases of question collocation:
\begin{equation}\label{eqn: congruity}
    \texttt{Congruity}(q_{s}, q_{t}) := \frac{1}{2}\left[\Delta(q_{s}, q_{t}) + \Delta(q_{t}, q_{s})\right]
\end{equation}
Although we have independently derived it from analyzing question collocations, our notion of question congruity coincides neatly with the established concept of \emph{pointwise mutual information} (PMI) between words~\cite{church-hanks-1989-word}, which has an equivalent mathematical form; by developing~\KCluster, we extend PMI to questions, which are more intricate than words.

\subsubsection{LLMs are exceptional probability machines}

Computing the PMI between words requires counting collocations; counting is, however, infeasible for calculating question congruity as two questions rarely, if at all, co-occur more than once in a collection of questions (e.g., a question almost never repeats itself in a well-designed exam). Instead, given a novel question pair, we need to extrapolate their collocation probabilities (in the form of $\log \Pr(q_{s} | q_{t})$ and $\log \Pr(q_{s})$) from existing data. LLMs, trained on massive corpora of diverse genres, are perfect for implementing question congruity because of their native ability to evaluate sophisticated text probabilities~\cite{jm3}. In this section, we describe an algorithm that uses Phi-2 to compute question congruity. 

% Our KC discovery algorithm~\KCluster~uses Phi-2's text evaluation ability to compute log-probabilities of the forms $\log \Pr(q_{s})$ and $\log \Pr(q_{s} | q_{t})$, so that it can aggregate congruent questions by their presumptive KCs.

% based on the idea of question congruity developed in Section~\ref{sec: congruity}. 

As an LLM, not only can Phi-2 extend a prompt (as in Section~\ref{sec: baseline}), but it can also evaluate the probability of alternative continuations to a given prompt. Let $\mathcal{P} := [p_{1}, p_{2}, \dots, p_{n}]$ denote a prompt comprising $n$ tokens ($p_{1}, \dots, p_{n}$) and $\mathcal{C} := [c_{1}, c_{2}, \dots, c_{k}]$ denote a \emph{prompt continuation} comprising $k$ tokens ($c_{1}, \dots, c_{k}$). To compute log-probabilities of the form $\log \Pr(q_{s} | q_{t})$, we consider $q_{t}$ as the prompt $\mathcal{P}$ and $q_{s}$ as a prompt continuation $\mathcal{C}$ to $\mathcal{P}$, and evaluate $\log\Pr(\mathcal{C} | \mathcal{P})$, the log-probability that $\mathcal{C}$ continues $\mathcal{P}$. The main algorithm is illustrated in Figure~\ref{fig: next-token}, along with three code snippets for executing each key step. 

The input to Phi-2 is a concatenation of the prompt and the continuation, $\mathcal{P} + \mathcal{C}$, producing an output of the same length. At each output location, Phi-2 generates a vector whose entries after a \texttt{log-softmax} normalization are the log-probability that each token in the vocabulary is to become the output token at that location, given the input tokens that Phi-2 has seen so far---in parti\-cular, one entry in the vector corresponds to the \emph{next token} in the input that Phi-2 has not consumed. For example, Figure~\ref{fig: next-token} shows that the output vector corresponding to the \emph{last token} $p_{n}$ in $\mathcal{P}$ contains an entry equal to $\log \Pr(c_{1} | \mathcal{P})$, the log-probability of the \emph{first token} $c_{1}$ in the continuation $\mathcal{C}$ conditioned on the entire prompt $\mathcal{P}$ that has been consumed before $c_{1}$ is; similarly, the output vector corresponding to the \emph{penultimate token} $c_{k - 1}$ in $\mathcal{C}$ contains an entry equal to $\log \Pr(c_{k} | \mathcal{C}_{<k}, \mathcal{P})$, the log-probability of the \emph{last token} $c_{k}$ in $\mathcal{C}$ conditioned on $\mathcal{P}$ and the partial continuation $\mathcal{C}_{<k}$ up to the $k$-th token. By the chain rule of probability, the target quantity $\log\Pr(\mathcal{C} | \mathcal{P})$ is simply the sum of these ``next-token'' log-probabilities starting from $\log\Pr(c_{1} | \mathcal{P})$, as shown at the bottom left of Figure~\ref{fig: next-token}, and can be calculated analogously to language modeling with a masked \texttt{cross-entropy} loss. 

% A token, more granular than a word, is the smallest linguistic element that Phi-2 can recognize.
% , and all tokens constitute Phi-2's \emph{vocabulary}. 
% We devised an algorithm that uses Phi-2 to compute $\log\Pr(\mathcal{C} | \mathcal{P})$, the log-probability of continuation $\mathcal{C}$ given the conditioning prompt $\mathcal{P}$; in particular, we think of $\mathcal{C}$ as $q_{s}$ and $\mathcal{P}$ as $q_{t}$, in connection with Equation~\ref{eqn: pmi} where we evaluate the conditional probability, $\log \Pr(q_{s} | q_{t})$. 

% Figure~\ref{fig: next-token} depicts the algorithm. 

% Phi-2 can efficiently process every token $c_{i}$ of $\mathcal{C}$ in parallel, and 

% \begin{equation}
%     \log \Pr(\mathcal{C} | \mathcal{P}) = \sum_{i} \log\Pr(c_{i} | \mathcal{C}_{<i}, \mathcal{P}) 
% \end{equation}

\begin{table}[t]
    \centering
    \caption{The prompt template used to evaluate $\log \Pr(q_{s} | q_{t})$ with a concrete example on the right. The template begins with $q_{t}$ (above the dashed line) as the conditioning prompt $\mathcal{P}$ and ends with $q_{s}$ (below the dashed line) as the prompt continuation $\mathcal{C}$, whose probability is to be evaluated.}
    \label{tab: kcluster_prompt}
\begin{adjustbox}{width=1.0\linewidth}
% {\setlength{\tabcolsep}{1.2em}
    \begin{tabular}
{p{0.48\linewidth}p{0.52\linewidth}}
    \toprule
    \vspace{-1.5em}
\begin{verbatim}
Exercise 1:
{question-type-1}:
{stem-1}
{
    choices-1
}
Answer: {answer-1} 

Exercise 2:
\end{verbatim}  & \vspace{-1.5em}
\begin{verbatim}
Exercise 1:
Multiple Choice:
Which is the most flexible?
a) bone
b) glass jar
c) rubber band
Answer: c) rubber band

Exercise 2:
\end{verbatim} \\  [-0.5em]
\hdashline
\vspace{-1em}
\begin{verbatim}
{question-type-2}:
{stem-2}
{
    choices-2
}
Answer: {answer-2}
\end{verbatim} & 
\vspace{-1em}
\begin{verbatim}
Multiple Choice:
Which is the most flexible?
a) paper
b) ceramic tea cup
c) clay tile
Answer: a) paper
\end{verbatim} \\ [-1em]
    \bottomrule
    \end{tabular}
\end{adjustbox}
\end{table}

To construct the input $\mathcal{P} + \mathcal{C}$ to evaluate $\log \Pr(q_{s} | q_{t})$ for two questions $q_{s}$ and $q_{t}$, we use the prompt template shown on the left of Table~\ref{tab: kcluster_prompt}. The template consists of two parts, separated by a dashed line. The upper part represents $\mathcal{P}$ in the algorithm, and sequentially contains the special marker ``\texttt{Exercise 1:}'' for introducing $q_{t}$, the content of $q_{t}$, and another special marker ``\texttt{Exercise 2:}'' for introducing $q_{s}$. The content of $q_{s}$, however, is contained in the lower part of the template, representing $\mathcal{C}$ in the algorithm. This design ensures that Phi-2 only evaluates the log-probability of $q_{s}$, while maintaining $q_{t}$ as the context. 

Calculating question congruity also requires computing mar\-ginal log-probabilities of the form $\log\Pr(q_{s})$, for which we use the same algorithm for computing $\log\Pr(\mathcal{C} | \mathcal{P})$ but keep $\mathcal{P}$ minimal. Table~\ref{tab: kcluster_prompt-single} shows the prompt template for computing the marginals, with a concrete example on the right. Compared to the prompt template in Table~\ref{tab: kcluster_prompt}, the new template removes all traces of the conditioning question $q_{t}$ from the upper part representing $\mathcal{P}$, but retains the special marker ``\texttt{Exercise 2:}'' for introducing $q_{s}$ in the lower part representing $\mathcal{C}$. This design ensures the algorithm closely approxi\-mates the genuine marginal log-probability $\log\Pr(q_{s})$ while keeping $\log\Pr(q_{s})$ compatible with $\log\Pr(q_{s} | q_{t})$ by only removing information about $q_{t}$.

Defined in terms of differences ($\Delta(q_{s}, q_{t})$ and $\Delta(q_{t}, q_{s})$), question congruity is invariant to the length of the questions, as the effect of length in $\log\Pr(q_{s} | q_{t})$ offsets that in $\log\Pr(q_{s})$, making it a versatile measure for different types and lengths of questions. Furthermore, question congruity captures more than text similarity, but an abstract notion of congruence (one question following another) that cosine-based metrics do not convey. We show in Section~\ref{sec: res_dis} that question congruity is more effective than negative cosine distance in measuring similar questions for clustering-based KC discovery. 

\begin{table}
    \centering
    \caption{The prompt template used to evaluate $\log \Pr(q_{s})$ with a concrete example on the right. The template uses the special marker ``\texttt{Exercise 2:}'' as the conditioning prompt $\mathcal{P}$ and the content of question $q_{s}$ (below the dashed line) as the prompt continuation $\mathcal{C}$, whose probability is to be evaluated.}
    \label{tab: kcluster_prompt-single}
\begin{adjustbox}{width=1.0\linewidth}
% {\setlength{\tabcolsep}{1.2em}
    \begin{tabular}
{p{0.48\linewidth}p{0.52\linewidth}}
    \toprule
    \vspace{-1.5em}
\begin{verbatim}
Exercise 2:
\end{verbatim}  & \vspace{-1.5em}
\begin{verbatim}
Exercise 2:
\end{verbatim} \\  [-0.5em]
\hdashline
\vspace{-1em}
\begin{verbatim}
{question-type-2}:
{stem-2}
{
    choices-2
}
Answer: {answer-2}
\end{verbatim} & 
\vspace{-1em}
\begin{verbatim}
Multiple Choice:
Which is the most flexible?
a) paper
b) ceramic tea cup
c) clay tile
Answer: a) paper
\end{verbatim} \\ [-1em]
    \bottomrule
    \end{tabular}
\end{adjustbox}
\end{table}

\section{Datasets}

We evaluate the four KC extraction methods described so far (\texttt{Concept}, \texttt{Concept-emb}, \texttt{Question-emb}, and \texttt{KCluster}) on three data\-sets of multiple-choice questions (MCQs) that vary in size and domain. All datasets include at least one expert-designed KC model that we consider as the gold standard in our evaluation, and two datasets contain additional data that allow us to validate each model on student transactions recorded in an actual class. 

\subsection{ScienceQA}

Based on various grade-school science curricula, \textbf{ScienceQA} \cite{lu2022learn} is a multi-modal dataset that covers three subjects: social science, language science, and natural science. Each question has two to four choices with one correct answer and comes with a ``skill'' tag---such as ``identify the experimental question''---that we consider as a KC label designed by an expert. To prepare the dataset for evaluation, we discarded questions accompanied by an image or tagged with a skill that appears less than ten times in all text-based questions, creating an evaluation subset of 10,701 MCQs.

\begin{table*}[t]
\centering
  \caption{KC alignment with \texttt{Skill} (99 KCs) assessed on ScienceQA}
  \label{tab: sciqa-quant}
  \begin{tabular}{lrcccccc}
    \toprule
      & & Adj. Rand & Adj. MI & FM Index & Homogeneity & Completeness & V-measure  \\
& & [-0.5, 1] & $(-\infty, 1]$ & [0, 1] & [0, 1] & [0, 1] & [0, 1]  \\
\midrule
\texttt{Concept} & (549 KCs) & 0.6454 & 0.8177 & 0.6603 & 0.9135 & 0.7925 & 0.8487 \\
\texttt{Question-emb} & (188 KCs) & \textbf{0.6940} & 0.8437 & \textbf{0.7036} & 0.9001 & 0.8308 & 0.8641 \\
\texttt{KCluster} & (198 KCs) & 0.6617 & \textbf{0.8513} & 0.6759 & \textbf{0.9157} & \textbf{0.8310} & \textbf{0.8713} \\
  \bottomrule
\end{tabular}
\end{table*}

\subsection{E-learning 2022}
Publicly available in DataShop~\cite{datashop}, the \textbf{E-learning 2022} data set\footnote{\url{https://pslcdatashop.web.cmu.edu/DatasetInfo?datasetId=5426}} contains questions and student activity data collected in a graduate e-learning design course taught between August and December 2022---a small subset of 80 MCQs were used in previous KC extraction work~\cite{moore_kc}. We parsed the course content in HTML documents and extracted 630 MCQs corresponding to 42,176 problem-solving attempts made by 39 students. In addition to the two default KC models, \texttt{Single-KC}, where all steps are labeled with a single KC, 
and \texttt{Unique-step}, where each step is labeled with a unique KC,
this dataset includes two expert-designed KC models based on learning objectives (LOs):
\texttt{LOs} and its improved version, \texttt{LOs-new}. In contrast to previous work~\cite{moore_kc}, we did not attempt to balance the number of MCQs per KC by curating a special subset of the MCQs, but retained the original mapping of MCQs to KCs in the expert-designed KC models for a more faithful evaluation of all methods.

\subsection{E-learning 2023} The \textbf{E-learning 2023} dataset\footnote{\url{https://pslcdatashop.web.cmu.edu/DatasetInfo?datasetId=5843}} is derived from the same e-learning course taught by a different instructor in a different semester (from August 2023 to December 2023). Unlike E-learning 2022, there was no course content available to extract questions from, so we chose 497 MCQs that are present in both years as the evaluation subset, which corresponds to 44,065 problem-solving attempts made by 41 students. 
This dataset also includes two expert-designed KC models: \texttt{v1-prompt-CTAmultimedia} (abbreviated as \texttt{v1-CTA}) and \texttt{v2-combined}, in addition to the two default KC models.

\section{Results and Discussion}
\label{sec: res_dis}

We use data to evaluate \texttt{KCluster} against three competing methods 
and answer our three RQs introduced earlier.
% in terms of our RQs about \textbf{label alignment}, \textbf{student mo\-deling}, and \textbf{KC improvement}.  

\subsection{Does KCluster align with expert-designed KC models? (RQ-1)}\label{sec: alignment}

\begin{table*}[t]
\centering
  \caption{KC alignment with \texttt{LOs-new} (101 KCs) assessed on E-learning 2022}
  \label{tab: elearning22-quant}
  \begin{tabular}{lrcccccc}
    \toprule
      & & Adj. Rand & Adj. MI & FM Index & Homogeneity & Completeness & V-measure  \\
& & [-0.5, 1] & $(-\infty, 1]$ & [0, 1] & [0, 1] & [0, 1] & [0, 1]  \\
\midrule
\texttt{Concept} & (371 KCs) & 0.2815 & 0.6337 & 0.3450 & \textbf{0.9328} & 0.7188 & \textbf{0.8119} \\
\texttt{Concept-emb} & (101 KCs) & 0.3090 & 0.6019 & 0.3220 & 0.7350 & 0.7240 & 0.7295 \\
\texttt{Question-emb} & (91 KCs) & 0.3533 & 0.6188 & 0.3668 & 0.7218 & 0.7439 & 0.7326 \\
\texttt{KCluster} & (114 KCs) & \textbf{0.4553} & \textbf{0.6939} & \textbf{0.4680} & 0.8139 & \textbf{0.7807} & 0.7970 \\
  \bottomrule
\end{tabular}
\end{table*}

\begin{table*}[t]
\centering
  \caption{KC alignment with \texttt{v1-CTA} (75 KCs) assessed on E-learning 2023}
  \label{tab: elearning23-quant}
  \begin{tabular}{lrcccccc}
    \toprule
      & & Adj. Rand & Adj. MI & FM Index & Homogeneity & Completeness & V-measure  \\
& & [-0.5, 1] & $(-\infty, 1]$ & [0, 1] & [0, 1] & [0, 1] & [0, 1]  \\
\midrule
\texttt{Concept} & (298 KCs) & 0.2888 & 0.6318 & 0.3511 & \textbf{0.9377} & 0.7092 & \textbf{0.8076} \\
\texttt{Concept-emb} & (81 KCs) & 0.3212 & 0.6091 & 0.3357 & 0.7384 & 0.7200 & 0.7291 \\
\texttt{Question-emb} & (78 KCs) & 0.3468 & 0.6385 & 0.3608 & 0.7535 & 0.7410 & 0.7472 \\
\texttt{KCluster} & (92 KCs) & \textbf{0.4361} & \textbf{0.7077} & \textbf{0.4529} & 0.8320 & \textbf{0.7776} & 0.8039 \\
  \bottomrule
\end{tabular}
\end{table*}

Although one can argue that no instructional expert could develop a flawless KC model and that expert opinions could diverge, alignment with expert-designed KC models provides quality assurance for automated KC extraction methods, as better alignment with human labels indicates more potential to be useful. 
% Our first research question is about assessing whether \texttt{KCluster} generates KC models that align with expert options and whether it does so better than other competing methods. 
In line with previous work~\cite{Matsuda_Wood_Shrivastava_Shimmei_Bier_2022}, we quantify the alignment of two KC models by comparing how they assign questions to KCs rather than counting text matches in KC labels---therefore, two models are perfectly aligned if both group the questions the same way, even if every group has a different label. Allowing different labels for the same KC reflects the multiple ways in which different experts can describe a KC and accounts for the nuances in different labeling approaches. Since a KC label indicates group membership analogously to a cluster label, \emph{regardless of whether a clustering algorithm is used}, we use standard metrics for \emph{clustering performance}\footnote{\url{https://scikit-learn.org/stable/modules/clustering.html\#clustering-performance-evaluation}} to assess how better \texttt{KCluster} aligns with expert-designed KC models than the other methods.

The following three metrics emphasize \emph{label agreement}: how well the predicted labels agree with the ground-truth classes. All methods are adjusted for chance, so that a random cluster assignment results in a score close to 0, whereas a perfect agreement has a score of 1:
\begin{itemize}[leftmargin=*]
    \item \textbf{Adjusted Rand Index} (Adj.\ Rand)~\cite{steinley_properties_2004}: a count-based measure popular in the literature; 
    \item \textbf{Adjusted Mutual Information} (Adj. MI)~\cite{10.1145/1553374.1553511}: an informa\-tion-theoretic measure adjusted for chance; 
    \item \textbf{Fowlkes-Mallows Index} (FM Index)~\cite{doi:10.1080/01621459.1983.10478008}: a measure based on pairwise precision and recall. 
\end{itemize}

The following three metrics highlight \emph{cluster quality}: how well each predicted cluster corresponds to the original classes. Low-quality assignments have a score close to 0 and perfect clusters have a score of 1, although a random assignment with a large number of clusters can have a specious, non-zero score (these three metrics are not adjusted for chance). 
\begin{itemize}[leftmargin=*]
    \item \textbf{Homogeneity}~\cite{rosenberg-hirschberg-2007-v}: a cluster assignment is homogeneous if every cluster contains only elements from the same ground-truth class;
    \item \textbf{Completeness}~\cite{rosenberg-hirschberg-2007-v}: a cluster assignment is complete if elements of the same ground-truth class are always assigned to the same cluster; 
     \item \textbf{V-measure}~\cite{rosenberg-hirschberg-2007-v}: the harmonic mean of homogeneity and completeness that balances both measures.  
\end{itemize}

Because no study has shown that one metric is more decisive than the others in assessing the alignment of KC models, we report all six metrics to give a more faithful evaluation of the four KC extraction methods. For all metrics, we use expert-designed KC labels as the gold standard, and if there is more than one expert-designed KC model, we choose the one that best fits student data as described in Section~\ref{sec: afm}. As no significant randomness is involved, we report the result of one execution of each method. 

\subsubsection{ScienceQA}

Table~\ref{tab: sciqa-quant} shows the results obtained from the ScienceQA data\-set, where the ``skill'' tag of each MCQ serves as ground-truth labels. With far fewer KCs (198 vs. 549), \texttt{KCluster} consistently outperforms \texttt{Concept}, the method based on extracting concepts from questions, in all six measures, showing closer alignment with the gold standard \texttt{Skill} model. \texttt{Question-emb}, based on question embeddings, also surpasses \texttt{Concept} in all metrics except homogeneity, running closely after \texttt{KCluster}. We excluded \texttt{Concept-emb}, the method based on concept embeddings, because it did not converge after 200 iterations of affinity propagation. 

The results on ScienceQA highlight that \texttt{Concept}, the most straightforward KC discovery method based on concept extraction using LLM, does not align with expert opinions better than the two clustering-based approaches, \texttt{KCluster} and \texttt{Question-emb}. Furthermore, \texttt{Concept} produces 4.5 times more KC labels than what is in the \texttt{Skill} model (549 vs. 99), which reaffirms the known limitation of this approach that it tends to produce excessive labels with word nuances. \texttt{KCluster}, however, generates an intermediate number of KCs and achieves the best score in four of the six metrics.

\subsubsection{E-learning 2022}
Table~\ref{tab: elearning22-quant} shows the results obtained from the E-learning 2022 data\-set, where \texttt{LOs-new}, the best expert-designed KC model according to Section~\ref{sec: elearning-22-afm}, serves as the gold standard. With an intermediate number of KCs, \texttt{KCluster} leads the other three methods on almost every metric, except that \texttt{Concept} has better homogeneity and V-measure scores. A high homogeneity score indicates that \texttt{Concept} has many KCs containing questions that belong to the same KC in the \texttt{LOs-new} model, but does not take into account whether questions belonging to the same KC in \texttt{LOs-new} are always assigned to the same KC in \texttt{Concept}---in fact, for questions belonging to the KC ``compare and contrast DFA and CTA skill'' in the \texttt{LOs-new} model, \texttt{Concept} created five KCs, two of which read ``a difficulty factors assessment'' and ``Difficulty Factors Assessment''. While \texttt{Concept} produced redundant labels as discussed previously, it also created the least \emph{complete} KC assignment where questions from the same ground-truth KC are scattered in multiple predicted KCs. In contrast, \texttt{KCluster} achieves the best completeness while maintaining the second-best homogeneity, despite marginally behind on the default V-measure that weighs both aspects equally. 

\subsubsection{E-learning 2023}
Table~\ref{tab: elearning23-quant} shows the results obtained from the E-learning 2023 dataset with \texttt{v1-CTA} as the gold standard. Since E-learning 2023 contains a subset of the questions in E-learning 2022, the results are consistent: \texttt{KCluster} outperforms all three other methods except that \texttt{Concept} has the best score in homogeneity and V-measure. \texttt{Concept} still created redundant KC labels and the least complete KC assignment---for a KC in \texttt{v1-CTA} about describing the redundancy principle in instructional design, \texttt{Concept} generated four KCs, three of which read ``redundancy'', ``redundancy principle'', and ``the redundancy principle''. To avoid redundant exposition, we conclude this section by highlighting that \textbf{with its lead on majority of the metrics, \texttt{KCluster} attained the best alignment with expert-designed KC models in all three datasets}.

\subsection{Does KCluster enable accurate prediction of student responses? (RQ-2)}\label{sec: afm}

While \texttt{KCluster}'s close alignment with expert-designed KC models suggests that \texttt{KCluster} is a promising approach, fit to student performance data provides a more reliable benchmark. An effective KC extraction method should produce an informative KC model (in the form of a binary Q-matrix~\cite{Barnes2005TheQM}) that an instructional expert can use with a statistical model to accurately predict student responses to questions. Our RQ-2 explores whether \texttt{KCluster} enables accurate student modeling, and if so, whether it outperforms the other methods. Using the student activity data from the E-learning 2022 and 2023 datasets, we train an Additive Factors Model (AFM)~\cite{afm} with the generated Q-matrices to evaluate the predictive power of each KC extraction method and report the standard metrics of model fit used by DataShop.

AFM~\cite{afm} is a logistic regression model that explains a student $i$'s correct (1) or incorrect (0) response to a question $j$ using the student's proficiency $\theta_{i}$ along with the KC difficulty $\beta_{k}$, the KC learning rate $\gamma_{k}$, and the number of student practices $T_{ik}$ for the relevant KCs as defined by a binary Q-matrix whose entry $q_{jk}$ indicates if question $j$ is associated with KC $k$. 
% AFM postulates that these factors produce an additive effect on student performance. 
If $Y_{ij}$ denotes a student $i$'s response to a question $j$, AFM computes the log-odds of the student giving correct response ($Y_{ij} = 1$) as a linear combination of these factors:
\begin{equation}\label{eq: afm}
    \log \frac{\Pr(Y_{ij} = 1)}{\Pr(Y_{ij} = 0)} = \theta_{i} + \sum_{k} q_{jk}\beta_{k} + q_{jk}\gamma_{k}T_{ik}
\end{equation}
Different KC extraction approaches tend to produce a different Q-matrix and thus instantiate a distinct AFM (via $q_{jk}$), for which the maximum likelihood estimation converges to different parameter estimates for $\theta_{i}$, $\beta_{k}$, and $\gamma_{k}$, allowing us to compare different approaches. Following the standard practice in DataShop\footnote{\url{https://pslcdatashop.web.cmu.edu/help?page=modelValues\#values}}, we report the Akaike information criterion (AIC) and Bayesian information criterion (BIC), which describe how well an AFM fits the \emph{current} data; moreover, we perform a cross-validation (CV) procedure that randomly divides questions (or \emph{items}) into \textbf{three folds} and repeat it with \textbf{50 different random seeds} to report the average item-stratified root mean square error (item-RMSE), which predicts how well an AFM generalizes to \emph{unseen} data. Stratifying the data by questions allows us to predict a student's responses to novel questions in the validation fold based on their responses to questions in the training folds, which is more relevant to our RQ-2. For all metrics, a lower value indicates a better prediction of student responses. 

% For these two datasets, we prefer item-blocked RMSE to student-blocked RMSE for measuring predictions on unseen data because, as the results show, the trivial \texttt{Unique-step} model achieves the best student-blocked RMSE in both data\-sets. This makes student-blocked RMSE unreliable as we would expect \texttt{Unique-step}, a model that assigns a unique KC to each item, to perform poorly on every metric (so that even worse models can be ``desk-rejected''). When \texttt{Unique-step} performs the best, however, we made the more logical decision to rely on other metrics and keep the dubious student-blocked RMSE only for the reader's reference. 

\begin{table}[t]
    \centering
    \caption{AFM performance on E-learning 22 (50 CV runs)}
    \label{tab: afm_22}
      \begin{adjustbox}{width=\linewidth}
    \begin{tabular}{lrccc}
        % \begin{tabularx}{\textwidth}{ lr|c c *{4}{Y} }
    \toprule
      & &  AIC & BIC  & Item-RMSE (Std.)  \\
     \midrule
     % \hline
\texttt{Single-KC} & (1 KC) &  46227.9805  & 46582.6144   & 0.4264 (0.0002) \\ 
\texttt{Unique-step} & (1,865 KCs) &  43323.0595  & 75923.4268  & 0.4273 (0.0002) \\ 
\midrule
\texttt{LOs} & (87 KCs) & 43972.6766  & 45815.0429  & 0.4244 (0.0010) \\ 
\texttt{LOs-new} & (101
 KCs) &  43353.2793  & \textbf{45437.8345}   & 0.4236 (0.0016) \\ 
\midrule
\texttt{Concept} & (371
 KCs) &  \textbf{41994.9029}  & 48750.2457  & 0.4295 (0.0017) \\ 
\texttt{Concept-emb} & (101 KCs) & 44537.1400  & 46621.6952  & 0.4292 (0.0011) \\ 
\texttt{Question-emb} & (91 KCs) &  43880.7030  & 45792.2660  & 0.4232 (0.0010) \\ 
\texttt{KCluster} & (114 KCs) & 43424.5571  & 45734.0021  & \textbf{0.4227} (0.0013) \\
\bottomrule
    \end{tabular}
% \end{tabularx}
\end{adjustbox}
\end{table}

\subsubsection{E-learning 2022}\label{sec: elearning-22-afm}
Table~\ref{tab: afm_22} summarizes the results on the E-learning 2022 data\-set. In addition to fitting an AFM with the Q-matrix generated by each automated KC extraction method, we also fit an AFM with the Q-matrix obtained from the two default KC models (\texttt{Single-KC} and \texttt{Unique-step}) and the two expert-designed models (\texttt{LOs} and \texttt{LOs-new}) for comparison. 
% ; it also has a decent student-blocked RMSE, but as we discussed, the trivial \texttt{Unique-step} model unexpectedly scores the best, making the results unreliable.

We observe that, although we did not use elaborate prompting strategies in our prompt template (Table~\ref{tab: baseline_prompt}), \texttt{Concept} is still a strong baseline with the best AIC among all models. The embedding-based approach, \texttt{Concept-emb}, managed to reduce the 371 KCs produced by \texttt{Concept} to 101 KCs via concept embedding and clustering, and consequently improved BIC, which favors models with fewer parameters. The other embedding-based approach \texttt{Question-emb}, however, outperforms \texttt{Concept-emb} in all metrics and achieves an item-RMSE comparable ($t(98) = -1.4738$, $p = .1437$) to that achieved by \texttt{LOs-new}, which has the best item-RMSE among expert KC models. This reinforces our initial prediction that encoding questions as embeddings should yield a better model than encoding concepts does, since questions contain more information than concepts do.

Using the novel question congruity to measure question simi\-larity, \texttt{KCluster} outperforms all other automated KC extraction methods except for having a higher AIC than \texttt{Concept}. In particular, \texttt{KCluster} significantly exceeds the best expert-designed KC model, \texttt{LOs-new}, in item-RMSE ($t(98) = -2.9963$, $p = .0035$) at $\alpha = .05$. Compared to \texttt{Question-emb}, which measures similar questions using the traditional negative cosine distance, \texttt{KCluster} fits to the student data better as evidenced by better AIC and BIC scores, and is likely to predict unseen data more accurately as evidenced by a better item-RMSE ($t(98) = -2.1145$, $p = .0370$). Together, these results suggest that it is advantageous to identify clusters of similar questions and assign KCs to clusters (as done by \texttt{KCluster}) rather than to individual questions (as done by \texttt{Concept}), and that question congruity is more effective than negative cosine distance for measuring similar questions in clustering-based KC discovery.

\subsubsection{E-learning 2023}
Table~\ref{tab: afm_23} shows the results obtained from the E-learning 2023 dataset, where we also trained an AFM for the two expert models, \texttt{v1-CTA} and \texttt{v2-combined}. Although all questions in E-learning 2023 are also present in E-learning 2022, the activity data come from a different student cohort, allowing us to assess whether each method is robust against different students. Consistent with what is observed in E-learning 2022, \texttt{KCluster} leads all three other automated methods in almost every metric, only slightly behind \texttt{Concept} on AIC; it has the best BIC score among all models, manual or automated, indicating that \texttt{KCluster} fits the current data the best. The two expert models have comparable scores on all measures, but \texttt{KCluster} outperforms both models in AIC and BIC, and significantly so in item-RMSE ($t(98) = -5.0956$, $p < .001$). In addition, \texttt{KCluster} significantly outperforms \texttt{Question-emb} in item-RMSE ($t(98) = -18.1487$, $p < .001$), reaffirming our conclusion from E-learning 2022 that \textbf{question congruity is superior to negative cosine distance in measuring question similarity}.

\begin{table}[t]
    \centering
    \caption{AFM performance on E-learning 23 (50 CV runs)}
    \label{tab: afm_23}
      \begin{adjustbox}{width=\linewidth}
    \begin{tabular}{lrccc}
        % \begin{tabularx}{\textwidth}{ lr|c c *{4}{Y} }
    \toprule
      & & AIC & BIC  & Item-RMSE (Std.)  \\
     \midrule
     % \hline
\texttt{Single-KC} & (1 KC) & 46210.3867  & 46566.8170   & 0.4141 (0.0001) \\
\texttt{Unique-step} & (1,398 KCs) &  42183.6839  & 66829.5327   & 0.4156 (0.0001) \\
\midrule
\texttt{v1-CTA} & (75 KCs) &  43434.4955  & 45077.5521   & 0.4088 (0.0021) \\
\texttt{v2-combined} & (72 KCs) &  43471.4342  & 45062.3302  & 0.4088 (0.0024) \\
\midrule
\texttt{Concept} & (298 KCs) & \textbf{41655.2518}  & 47175.5742  & 0.4111 (0.0021) \\
\texttt{Concept-emb} & (81 KCs) &  44366.9480  & 46114.3256  & 0.4151 (0.0014) \\
\texttt{Question-emb} & (78 KCs) &   43946.2607  & 45641.4778  & 0.4108 (0.0011) \\
\texttt{KCluster} & (92 KCs) &  42999.9064  & \textbf{44938.5393}  & \textbf{0.4071} (0.0009) \\
\bottomrule
    \end{tabular}
% \end{tabularx}
\end{adjustbox}
\end{table}

\begin{table}[t]
    \centering
    \caption{Improvements to the KC ``11.1 apply\_evidence'' in E-learning 22 (50 CV runs)}
    \label{tab: imp}
    \begin{adjustbox}{width=\linewidth}
    \begin{tabular}{lrccc}
    \toprule
      & & AIC & BIC  & Item-RMSE (Std.)  \\
     \midrule
     % \hline
\texttt{LOs-new} & & 43353.2716  & 45437.8268  & 0.4230 (0.0012) \\
\texttt{Concept} & ($+5$ KCs) & 43265.7209  & 45419.4730  & 0.4223 (0.0012) \\
\texttt{Question-emb} & ($+3$ KCs)  & 43284.1397  & 45403.2933  & 0.4225 (0.0012) \\
\texttt{KCluster} & ($+4$ KCs) & \textbf{43262.7614}  & \textbf{45399.2143}  & \textbf{0.4221} (0.0013) \\
\bottomrule
    \end{tabular}
    \end{adjustbox}
\end{table}

% \begin{table}[t]
%     \centering
%     \caption{Comparison of KC breakdowns}
%     \label{tab: kc-breakdown}
%       \begin{adjustbox}{width=\linewidth}
%     \begin{tabular}{ccc}
%     \toprule
%  KC (\texttt{Concept}) & KC (\texttt{Question-emb}) & KC (\texttt{KCluster}) \\
% \midrule
%  generative processing & \textbf{generative and extraneous processing} & \textbf{generative and extraneous processing} \\
%  \textbf{evidence} & the practice or testing effect & \textbf{evidence} \\
%  \textbf{evidence} & the practice or testing effect & \textbf{evidence} \\
%  \textbf{evidence} & the practice or testing effect & \textbf{evidence} \\
% \bottomrule
%     \end{tabular}
%     \end{adjustbox}
% \end{table}

% \begin{table*}[t]
%     \centering
%     \caption{Comparison of KC breakdowns}
%     \label{tab: kc-breakdown}
%       % \begin{adjustbox}{width=\textwidth}
%     \begin{tabular}{cccc}
%     \toprule
% KC (\texttt{LOs-new}) & KC (\texttt{Concept}) & KC (\texttt{Question-emb}) & KC (\texttt{KCluster}) \\
% \midrule
% 11.1 apply\_evidence & generative processing & \textbf{generative and extraneous processing} & \textbf{generative and extraneous processing} \\
% 11.1 apply\_evidence & \textbf{evidence} & the practice or testing effect & \textbf{evidence} \\
% 11.1 apply\_evidence & \textbf{evidence} & the practice or testing effect & \textbf{evidence} \\
% 11.1 apply\_evidence & \textbf{evidence} & the practice or testing effect & \textbf{evidence} \\
% \bottomrule
%     \end{tabular}
% \end{table*}

\begin{figure*}
    \centering
    \includegraphics[width=\textwidth]{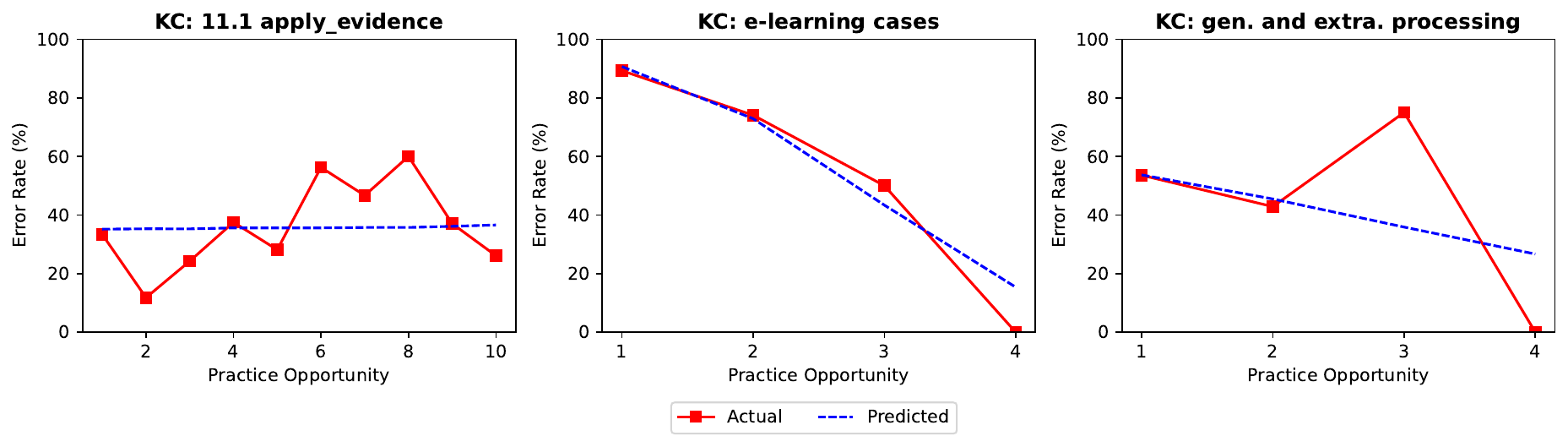}
    \caption{The learning curves for the original expert KC (left) and two new KCs discovered by \texttt{KCluster} (middle and right)}
    \label{fig: lc}
    \Description{This figure shows the learning curve of the original KC ``apply evidence'' on the left, and the learning curves of the two new KCs produced by KCluster on the right.}
\end{figure*}

\subsection{Does KCluster reveal insights about problematic KCs? (RQ-3)} \label{sec: insights}
By generating an alternate KC model, \texttt{KCluster} suggested how questions could have been organized by KCs so that an instructor can better predict student responses, but it did not explain, for example, why learning was difficult for some problematic KCs in the original expert KC model. A KC is problematic (and worth investigating) if it is neither too difficult nor too easy to learn, yet the students did not show any learning~\cite{stamper_kc}. Previous work using data-driven DFA~\cite{stamper_kc} manually analyzed and divided a problematic KC into three hidden KCs, which improved the prediction of student responses when reinserted into the original model. Our RQ-3 explores how \texttt{KCluster} can \emph{automatically} reveal similar insights about and suggest improvements to problematic KCs.

% An effective automated KC extraction method should also illuminate improvements to the current curriculum. 

From the E-learning 2022 dataset, we first identified 14 problematic KCs in the \texttt{LOs-new} model that an AFM estimated to have a learning rate $\gamma_{k} < 0.001$ (students were not learning) and an initial success probability (equal to $\texttt{sigmoid}(\beta_{k})$) between 0.2 and 0.8 (the KC was neither too difficult nor too easy to learn). Following previous work~\cite{stamper_kc}, we then applied \texttt{Concept}, \texttt{Question-emb}, and \texttt{KCluster} to the questions associated with each problematic KC and discovered new KCs that constitute the original. We searched for improvement, where an AFM achieves a lower item-RMSE, in each new KC model that had a problematic KC replaced, and found that \texttt{Concept} and \texttt{KCluster} significantly improved the KC ``11.1 apply\_evidence'', which has a zero learning rate and an initial success probability of 0.65. 

Table~\ref{tab: imp} quantifies the improvements. Compared to \texttt{LOs-new}, all methods divide the original KC into multiple new KCs, suggesting that the expert KC is too coarse to reflect a single skill. \texttt{KCluster} breaks the generic ``11.1 apply\_evidence'' KC into four different KCs, three of which concern specific sources of evidence (``generative and extraneous processing'', ``the practice or testing effect'', and ``e-learning cases''), with a fourth ``evidence'' KC for problems that contrast evidence and ask students to decide which situation would yield better learning. When reinserted into the original \texttt{LOs-new} model, the four new KCs discovered by \texttt{KCluster} brought the greatest improvements in all three metrics and significantly in item-RMSE ($t(98) = -3.4379, p < .001$).

% searched for improvements to those KCs in \texttt{LOs-new} that have a learning rate less than $0.001$ and an initial success probability between 0.2 and 0.8, and 

% Looking at the new KC mappings (Table 11), it is evident that, compared to the expert-developed KC (LOs-new), all generated KCs divide the steps in more than one KC, suggesting that the expert-generated KC is over-general and unlikely to reflect a single skill. 

% KCluster divides the original KC into 4 different KCs. Looking at the KC mappings (Table~\ref{tab: kc-breakdown}), we see that instead of a generic ``Apply Evidence'' KC, these data are better captured by a series of KCs focusing on the specific source of evidence (generative and extraneous processing, the practice or testing effect, and e-learning cases), leaving a fourth KC for problems that contrast evidence and ask students to decide which of two situations would yield best learning (see Table 10 for an example two such KCs).

This automated DFA not only discovers more specific and potentially more meaningful KCs, but also captures student learning better. Figure~\ref{fig: lc} contrasts the learning curve of the original ``11.1 apply\_evidence'' KC with that of two new KCs (``e-learning cases'' and ``generative and extraneous processing'') created by \texttt{KCluster}. While the original learning curve remains flat after ten learning opportunities, the err\-or rates depicted in the new learning curves quickly approach zero after four opportunities, showing clear evidence of learning. An instructor, after reviewing the new learning curves, will be able to make informed adjustments to instruction and improve student learning \emph{specifically} in the other two aspects of ``applying evidence'', with which students were struggling (namely, ``the practice or testing effect'' and ``evidence''). This shows that \textbf{\texttt{KCluster} is not only capable of predicting student responses in foresight, but it can also illuminate improvements to instruction in retrospect}.

\section{General Discussion}
Our comprehensive evaluation reveals three critical insights about \texttt{KCluster} that we will discuss in this section.

\textbf{Clustering-based approaches outperform concept extraction}. Using the text generation ability of Phi-2, \texttt{Concept} is a natural LLM-based method to extract KCs from questions. Yet, \emph{using the same LLM}, \texttt{KCluster} shows that closer alignment with expert models (Section~\ref{sec: alignment}), better prediction of student responses (Section~\ref{sec: afm}), and greater improvement to problematic KCs (Section~\ref{sec: insights}) can be achieved by coupling Phi-2's native ability to evaluate text probabilities with clustering. That we chose Phi-2 over more advanced LLMs for Phi-2's balanced performance and affordability does not account for this performance discrepancy, as both methods use the same LLM. In fact, using a more advanced LLM and a curated set of 80 MCQs from the E-learning 2022 dataset, previous work~\cite{moore_kc} only managed to produce the exact KC for 28 MCQs (35\%). A possible reason for this low KC match rate is that the powerful LLM generated redundant labels with undesired word nuances. Clustering-based approaches like \texttt{KCluster}, on the other hand, reduce the redundancy by propagating the labels of the cluster exemplars. As a rising tide will lift all boats, we expect future work using a more advanced LLM to improve both classes of methods, but Phi-2 is free and therefore more readily available to instructors.
% preferably not at the cost of accessibility for instructors.

\textbf{Question congruity is more effective than negative cosine distance in measuring similar questions for clustering-based KC discovery.} Both \texttt{KCluster} and \texttt{Question-emb} use affinity propagation~\cite{doi:10.1126/science.1136800} to identify clusters of similar questions and label all questions in a cluster with a KC equivalent to the concept label of the cluster exemplar. \texttt{KCluster}, however, outperforms \texttt{Question-emb} in aligning with expert-designed KC models, predicting student responses, and improving problematic KCs, by using the novel question congruity described in Section~\ref{sec: kcluster} (rather than the traditional negative cosine distance) to measure question similarity. These posi\-tive results have strengthened our belief that future work will prove question congruity a strong measure of question similarity in more domains than KC discovery. 

% \textbf{Clustering-based approaches outperform concept extraction}. Using the text generation ability of Phi-2, \texttt{Concept} is a natural LLM-based method to extract KCs from questions. Yet, \emph{using the same LLM}, \texttt{KCluster} shows that closer alignment with expert models, better prediction of student responses, and greater improvement to problematic KCs can be achieved instead by coupling Phi-2's native ability to evaluate text probabilities with clustering. That we chose Phi-2 over more advanced LLMs for Phi-2's balanced performance and affordability does not account for this performance discrepancy, as both methods use the same LLM; in fact, using a more advanced LLM and a curated set of 80 MCQs from the E-learning 2022 dataset, previous work~\cite{moore_kc} only managed to produce the exact KC for 28 MCQs (35\%), a notable reason for which is that the powerful LLM generated redundant labels with undesired word nuances. Clustering-based approaches like \texttt{KCluster} reduce the redundancy by propagating the labels of the cluster exemplars. As a rising tide will lift all boats, we expect future work using a more advanced LLM to improve both classes of methods, preferably not at the cost of accessibility for instructors.

\textbf{Automated approaches can outperform manual approaches.} Combining the strengths of LLM and clustering, \texttt{KCluster} enables instructors to predict student responses better than the best expert model does in the two e-learning datasets (Section~\ref{sec: afm}). While we expect future work to extend \texttt{KCluster} to more datasets and more question types, our evaluation offers strong evidence that \texttt{KCluster}, an automated approach, can surpass manual approaches in modeling student learning. Furthermore, \texttt{KCluster} has demonstrated initial success in automated DFA (Section~\ref{sec: insights}), inspiring future work that closes the loop by implementing and validating new instructional designs informed by \texttt{KCluster}.

\section{Conclusion}

We proposed question congruity, a novel measure of question similarity based on question collocations, and described an algorithm that uses Phi-2 to compute the required probabilities. The two contributions underlie \texttt{KCluster}, a novel KC discovery approach that combines LLM and clustering. Our comprehensive evaluation shows that \texttt{KCluster} not only outperforms the other three competing methods and the best expert KC model, but can also offer insights into problematic KCs that potentially inspire new instructional designs.

% \begin{table}[t]
% \centering
%   \caption{Comparison of KC discovery approaches}
%   \label{tab: related-work}
%     \begin{tabularx}{\textwidth}{c*{1}{Y} | c*{4}{Y} }
%      \toprule
%     \textbf{Method} & \textbf{Input} &  & Obviates Human Input & Produces Descriptive KCs & Is Validated on Student Data \\ 
%     % \textbf{Method} & \textbf{Input}  & &  & &  \\
%     \hline
%     CTA~\cite{Clark2008} & Questions & & \xmark & \cmark & \cmark \\
%     Crowdsourcing & & & \xmark & \cmark & \cmark \\
%     \hline
%     LFA~\cite{lfa} &  Difficulty Factors & & \xmark & \cmark & \cmark \\
%     \hline
%     Q-matrix~\cite{Barnes2005TheQM} & Student Data & & \xmark & \xmark & \cmark \\
% Factorization~\cite{10.1007/978-3-642-30950-2_58,10.1007/978-3-642-39112-5_45,JMLR:v15:lan14a} & Student Data & & \xmark & \xmark & \cmark \\
%     Statistical 
%  & Student & & \xmark & \xmark & \cmark \\
%     Learning~\cite{liu_data-driven_2012,pardos_dafm_2018,2022.EDM-long-papers.2,GonzlezBrenes2013WhatAW,shi_kc-finder_2023}
%  & Data & &  &  &  \\
%     GPT-4~\cite{moore_kc} & Questions & & \cmark & \cmark & \xmark \\
%     \textbf{KCluster (Ours)} & Questions & & \cmark & \cmark & \cmark \\
%   \bottomrule
% \end{tabularx}
% \end{table}

%ACKNOWLEDGMENTS are optional
\section{Acknowledgments}
This research was supported by the National Science Foundation under Grant No. 2301130 and a Google Academic Research Award to Paulo F. Carvalho.

This research was also supported by a US Navy grant on Real-time Knowledge Sharing awarded to John Stamper under Grant No. N68335-23-C-0035.   

% Funding to attend this conference was provided by the CMU GSA/Provost Conference Funding.

% to acknowledge grants, funding, editing assistance and
% what have you.  In the present case, for example, the
% authors would like to thank Gerald Murray of ACM for
% his help in codifying this \textit{Author's Guide}
% and the \textbf{.cls} and \textbf{.tex} files that it describes. Acknowledgments should be left blank during the review process.

% \balancecolumns

% \clearpage

\bibliographystyle{abbrv}
\bibliography{mybibliography}

\begin{thebibliography}{10}

\bibitem{Barnes2005TheQM}
T.~Barnes.
\newblock The q-matrix method: Mining student response data for knowledge.
\newblock In {\em AAAI Workshop}, 2005.

\bibitem{lfa}
H.~Cen, K.~Koedinger, and B.~Junker.
\newblock Learning factors analysis -- a general method for cognitive model evaluation and improvement.
\newblock In M.~Ikeda, K.~D. Ashley, and T.-W. Chan, editors, {\em Intelligent Tutoring Systems}, pages 164--175, Berlin, Heidelberg, 2006. Springer Berlin Heidelberg.

\bibitem{afm}
H.~Cen, K.~R. Koedinger, and B.~Junker.
\newblock Is over practice necessary? improving learning efficiency with the cognitive tutor through educational data mining.
\newblock In {\em Proceedings of the 2007 Conference on Artificial Intelligence in Education: Building Technology Rich Learning Contexts That Work}, page 511–518, NLD, 2007. IOS Press.

\bibitem{chau_automatic_2021}
H.~Chau, I.~Labutov, K.~Thaker, D.~He, and P.~Brusilovsky.
\newblock Automatic {Concept} {Extraction} for {Domain} and {Student} {Modeling} in {Adaptive} {Textbooks}.
\newblock {\em International Journal of Artificial Intelligence in Education}, 31(4):820--846, Dec. 2021.

\bibitem{chen2021evaluating}
M.~Chen, J.~Tworek, H.~Jun, Q.~Yuan, H.~P. d.~O. Pinto, J.~Kaplan, H.~Edwards, Y.~Burda, N.~Joseph, G.~Brockman, A.~Ray, R.~Puri, G.~Krueger, M.~Petrov, H.~Khlaaf, G.~Sastry, P.~Mishkin, B.~Chan, S.~Gray, N.~Ryder, M.~Pavlov, A.~Power, L.~Kaiser, M.~Bavarian, C.~Winter, P.~Tillet, F.~P. Such, D.~Cummings, M.~Plappert, F.~Chantzis, E.~Barnes, A.~Herbert-Voss, W.~H. Guss, A.~Nichol, A.~Paino, N.~Tezak, J.~Tang, I.~Babuschkin, S.~Balaji, S.~Jain, W.~Saunders, C.~Hesse, A.~N. Carr, J.~Leike, J.~Achiam, V.~Misra, E.~Morikawa, A.~Radford, M.~Knight, M.~Brundage, M.~Murati, K.~Mayer, P.~Welinder, B.~McGrew, D.~Amodei, S.~McCandlish, I.~Sutskever, and W.~Zaremba.
\newblock Evaluating large language models trained on code.
\newblock {\em arXiv preprint arXiv:2107.03374}, July 2021.

\bibitem{church-hanks-1989-word}
K.~W. Church and P.~Hanks.
\newblock Word association norms, mutual information, and lexicography.
\newblock In {\em 27th Annual Meeting of the Association for Computational Linguistics}, pages 76--83, Vancouver, British Columbia, Canada, June 1989. Association for Computational Linguistics.

\bibitem{Clark2008}
R.~E. Clark, D.~Feldon, J.~J.~G. van Merriënboer, K.~Yates, and S.~Early.
\newblock Cognitive task analysis.
\newblock In J.~M. Spector, M.~D. Merrill, J.~J.~G. van Merriënboer, and M.~P. Driscoll, editors, {\em Handbook of research on educational communications and technology}, pages 577--593. Macmillan/Gale, New York, 3rd edition, 2008.

\bibitem{cobbe2021gsm8k}
K.~Cobbe, V.~Kosaraju, M.~Bavarian, M.~Chen, H.~Jun, L.~Kaiser, M.~Plappert, J.~Tworek, J.~Hilton, R.~Nakano, C.~Hesse, and J.~Schulman.
\newblock Training verifiers to solve math word problems.
\newblock {\em arXiv preprint arXiv:2110.14168}, October 2021.

\bibitem{corbett_knowledge_1994}
A.~T. Corbett and J.~R. Anderson.
\newblock Knowledge tracing: Modeling the acquisition of procedural knowledge.
\newblock {\em User Modeling and User-Adapted Interaction}, 4(4):253--278, Dec. 1994.

\bibitem{dina}
J.~de~la Torre.
\newblock Dina model and parameter estimation: A didactic.
\newblock {\em Journal of Educational and Behavioral Statistics}, 34(1):115--130, 2009.

\bibitem{10.1007/978-3-642-30950-2_58}
M.~C. Desmarais, B.~Beheshti, and R.~Naceur.
\newblock Item to skills mapping: Deriving a conjunctive q-matrix from data.
\newblock In S.~A. Cerri, W.~J. Clancey, G.~Papadourakis, and K.~Panourgia, editors, {\em Intelligent Tutoring Systems}, pages 454--463, Berlin, Heidelberg, 2012. Springer Berlin Heidelberg.

\bibitem{10.1007/978-3-642-39112-5_45}
M.~C. Desmarais and R.~Naceur.
\newblock A matrix factorization method for mapping items to skills and for enhancing expert-based q-matrices.
\newblock In H.~C. Lane, K.~Yacef, J.~Mostow, and P.~Pavlik, editors, {\em Artificial Intelligence in Education}, pages 441--450, Berlin, Heidelberg, 2013. Springer Berlin Heidelberg.

\bibitem{devlin2019bert}
J.~Devlin, M.-W. Chang, K.~Lee, and K.~Toutanova.
\newblock Bert: Pre-training of deep bidirectional transformers for language understanding.
\newblock {\em Proceedings of the 2019 Conference of the North American Chapter of the Association for Computational Linguistics: Human Language Technologies, Volume 1 (Long and Short Papers)}, pages 4171--4186, June 2019.

\bibitem{doi:10.1080/01621459.1983.10478008}
E.~B. Fowlkes and C.~L. Mallows.
\newblock A method for comparing two hierarchical clusterings.
\newblock {\em Journal of the American Statistical Association}, 78(383):553--569, 1983.

\bibitem{doi:10.1126/science.1136800}
B.~J. Frey and D.~Dueck.
\newblock Clustering by passing messages between data points.
\newblock {\em Science}, 315(5814):972--976, 2007.

\bibitem{GonzlezBrenes2013WhatAW}
J.~P. Gonz{\'a}lez-Brenes and J.~Mostow.
\newblock What and when do students learn? fully data-driven joint estimation of cognitive and student models.
\newblock In {\em Educational Data Mining}, 2013.

\bibitem{gunasekar2023textbooks}
S.~Gunasekar, Y.~Zhang, J.~Aneja, C.~C.~T. Mendes, A.~D. Giorno, S.~Gopi, M.~Javaheripi, P.~Kauffmann, G.~de~Rosa, O.~Saarikivi, A.~Salim, S.~Shah, H.~S. Behl, X.~Wang, S.~Bubeck, R.~Eldan, A.~T. Kalai, Y.~T. Lee, and Y.~Li.
\newblock Textbooks are all you need, 2023.

\bibitem{HambletonSwaminathan1985}
R.~Hambleton and H.~Swaminathan.
\newblock {\em Item Response Theory: Principles and Applications}.
\newblock Springer Science+Business Media, New York, NY, USA, 1985.

\bibitem{HeffernanKoedinger1998}
N.~T. Heffernan and K.~R. Koedinger.
\newblock A developmental model for algebra symbolization: The results of a difficulty factors assessment.
\newblock In {\em Proceedings of the Twentieth Annual Conference of the Cognitive Science Society}, pages 484--489, Mahwah, NJ, 1998. Lawrence Erlbaum Associates, Inc.

\bibitem{hughes_phi-2_2023}
M.~Javaheripi, S.~Bubeck, M.~Abdin, J.~Aneja, S.~Bubeck, C.~C.~T. Mendes, W.~Chen, A.~D. Giorno, R.~Eldan, S.~Gopi, S.~Gunasekar, M.~Javaheripi, P.~Kauffmann, Y.~T. Lee, Y.~Li, A.~Nguyen, G.~de~Rosa, O.~Saarikivi, A.~Salim, S.~Shah, M.~Santacroce, H.~S. Behl, A.~T. Kalai, X.~Wang, R.~Ward, P.~Witte, C.~Zhang, and Y.~Zhang.
\newblock Phi-2: {The} surprising power of small language models, Dec. 2023.

\bibitem{jiang2023mistral}
A.~Q. Jiang, A.~Sablayrolles, A.~Mensch, C.~Bamford, D.~S. Chaplot, D.~d.~l. Casas, F.~Bressand, G.~Lengyel, G.~Lample, L.~Saulnier, T.~Lavril, M.-A. Lachaux, D.~Massiceti, J.~Rio, R.~Lambert, S.~Bhosale, S.~Aminov, W.~Kool, R.~Everett, A.~Gu, S.~Dukma, H.~Hao, X.~Zhou, J.~Chen, C.~Iovine, W.~Chen, V.~Wang, and J.~Calandriello.
\newblock Mistral 7b.
\newblock {\em arXiv preprint arXiv:2310.06825}, October 2023.

\bibitem{jm3}
D.~Jurafsky and J.~H. Martin.
\newblock {\em Speech and Language Processing: An Introduction to Natural Language Processing, Computational Linguistics, and Speech Recognition with Language Models}.
\newblock 3rd edition, 2024.
\newblock Online manuscript released August 20, 2024.

\bibitem{kli}
K.~R. Koedinger, A.~T. Corbett, and C.~Perfetti.
\newblock The knowledge-learning-instruction framework: Bridging the science-practice chasm to enhance robust student learning.
\newblock {\em Cognitive Science}, 36(5):757--798, 2012.

\bibitem{Koedinger2010-KOESLL}
K.~R. Koedinger and E.~A. McLaughlin.
\newblock Seeing language learning inside the math: Cognitive analysis yields transfer.
\newblock In S.~Ohlsson and R.~Catrambone, editors, {\em Proceedings of the 32nd Annual Conference of the Cognitive Science Society}, pages 471--476. Cognitive Science Society, 2010.

\bibitem{koedinger_automated_2012}
K.~R. Koedinger, E.~A. McLaughlin, and J.~C. Stamper.
\newblock Automated {Student} {Model} {Improvement}.
\newblock Technical report, International Educational Data Mining Society, June 2012.
\newblock ERIC Number: ED537201.

\bibitem{Koedinger2004}
K.~R. Koedinger and M.~J. Nathan.
\newblock The real story behind story problems: Effects of representations on quantitative reasoning.
\newblock {\em Journal of the Learning Sciences}, 13(2):129--164, 2004.

\bibitem{10.1007/978-3-642-39112-5_43}
K.~R. Koedinger, J.~C. Stamper, E.~A. McLaughlin, and T.~Nixon.
\newblock Using data-driven discovery of better student models to improve student learning.
\newblock In H.~C. Lane, K.~Yacef, J.~Mostow, and P.~Pavlik, editors, {\em Artificial Intelligence in Education}, pages 421--430, Berlin, Heidelberg, 2013. Springer Berlin Heidelberg.

\bibitem{koedinger_testing_2016}
K.~R. Koedinger, M.~V. Yudelson, and P.~I. Pavlik~Jr.
\newblock Testing {Theories} of {Transfer} {Using} {Error} {Rate} {Learning} {Curves}.
\newblock {\em Topics in Cognitive Science}, 8(3):589--609, 2016.

\bibitem{JMLR:v15:lan14a}
A.~S. Lan, A.~E. Waters, C.~Studer, and R.~G. Baraniuk.
\newblock Sparse factor analysis for learning and content analytics.
\newblock {\em Journal of Machine Learning Research}, 15(57):1959--2008, 2014.

\bibitem{DBLP:conf/edm/LiCKM11}
N.~Li, W.~W. Cohen, K.~R. Koedinger, and N.~Matsuda.
\newblock A machine learning approach for automatic student model discovery.
\newblock In M.~Pechenizkiy, T.~Calders, C.~Conati, S.~Ventura, C.~Romero, and J.~C. Stamper, editors, {\em Proceedings of the 4th International Conference on Educational Data Mining, Eindhoven, The Netherlands, July 6-8, 2011}, pages 31--40. www.educationaldatamining.org, 2011.

\bibitem{li_general_2013}
N.~Li, E.~Stampfer, W.~Cohen, and K.~Koedinger.
\newblock General and {Efficient} {Cognitive} {Model} {Discovery} {Using} a {Simulated} {Student}.
\newblock {\em Proceedings of the Annual Meeting of the Cognitive Science Society}, 35(35), 2013.

\bibitem{liu_data-driven_2012}
J.~Liu, G.~Xu, and Z.~Ying.
\newblock Data-{Driven} {Learning} of {Q}-{Matrix}.
\newblock {\em Applied psychological measurement}, 36(7):548--564, Oct. 2012.

\bibitem{Liu_Koedinger_2017}
R.~Liu and K.~R. Koedinger.
\newblock Closing the loop: Automated data-driven cognitive model discoveries lead to improved instruction and learning gains.
\newblock {\em Journal of Educational Data Mining}, 9(1):25–41, Sep. 2017.

\bibitem{lu2022learn}
P.~Lu, S.~Mishra, T.~Xia, L.~Qiu, K.-W. Chang, S.-C. Zhu, O.~Tafjord, P.~Clark, and A.~Kalyan.
\newblock Learn to explain: Multimodal reasoning via thought chains for science question answering.
\newblock In {\em The 36th Conference on Neural Information Processing Systems (NeurIPS)}, 2022.

\bibitem{Matsuda_Wood_Shrivastava_Shimmei_Bier_2022}
N.~Matsuda, J.~Wood, R.~Shrivastava, M.~Shimmei, and N.~Bier.
\newblock Latent skill mining and labeling from courseware content.
\newblock {\em Journal of Educational Data Mining}, 14(2), Oct. 2022.

\bibitem{mihalcea-tarau-2004-textrank}
R.~Mihalcea and P.~Tarau.
\newblock {T}ext{R}ank: Bringing order into text.
\newblock In D.~Lin and D.~Wu, editors, {\em Proceedings of the 2004 Conference on Empirical Methods in Natural Language Processing}, pages 404--411, Barcelona, Spain, July 2004. Association for Computational Linguistics.

\bibitem{moore_kc}
S.~Moore, R.~Schmucker, T.~Mitchell, and J.~Stamper.
\newblock Automated generation and tagging of knowledge components from multiple-choice questions.
\newblock In {\em Proceedings of the Eleventh ACM Conference on Learning @ Scale}, L@S '24, page 122–133, New York, NY, USA, 2024. Association for Computing Machinery.

\bibitem{2022.EDM-long-papers.2}
B.~PaaÃŸen, M.~Dywel, M.~Fleckenstein, and N.~Pinkwart.
\newblock Sparse factor autoencoders for item response theory.
\newblock In A.~Mitrovic and N.~Bosch, editors, {\em Proceedings of the 15th International Conference on Educational Data Mining}, pages 17--26, Durham, United Kingdom, July 2022. International Educational Data Mining Society.

\bibitem{pardos_dafm_2018}
Z.~A. Pardos and A.~Dadu.
\newblock {dAFM}: {Fusing} {Psychometric} and {Connectionist} {Modeling} for {Q}-matrix {Refinement}.
\newblock {\em Journal of Educational Data Mining}, 10(2):1--27, Oct. 2018.
\newblock Number: 2.

\bibitem{10.5555/3454287.3455008}
A.~Paszke, S.~Gross, F.~Massa, A.~Lerer, J.~Bradbury, G.~Chanan, T.~Killeen, Z.~Lin, N.~Gimelshein, L.~Antiga, A.~Desmaison, A.~K\"{o}pf, E.~Yang, Z.~DeVito, M.~Raison, A.~Tejani, S.~Chilamkurthy, B.~Steiner, L.~Fang, J.~Bai, and S.~Chintala.
\newblock {\em PyTorch: an imperative style, high-performance deep learning library}.
\newblock Curran Associates Inc., Red Hook, NY, USA, 2019.

\bibitem{reimers-2019-sentence-bert}
N.~Reimers and I.~Gurevych.
\newblock Sentence-bert: Sentence embeddings using siamese bert-networks.
\newblock In {\em Proceedings of the 2019 Conference on Empirical Methods in Natural Language Processing}. Association for Computational Linguistics, 11 2019.

\bibitem{rosenberg-hirschberg-2007-v}
A.~Rosenberg and J.~Hirschberg.
\newblock {V}-measure: A conditional entropy-based external cluster evaluation measure.
\newblock In J.~Eisner, editor, {\em Proceedings of the 2007 Joint Conference on Empirical Methods in Natural Language Processing and Computational Natural Language Learning ({EMNLP}-{C}o{NLL})}, pages 410--420, Prague, Czech Republic, June 2007. Association for Computational Linguistics.

\bibitem{RussellNorvig2020}
S.~J. Russell and P.~Norvig.
\newblock {\em Artificial Intelligence: A Modern Approach}.
\newblock Pearson, 4th edition, 2020.

\bibitem{shabana_unsupervised_2023}
K.~M. Shabana and C.~Lakshminarayanan.
\newblock Unsupervised {Concept} {Tagging} of {Mathematical} {Questions} from {Student} {Explanations}.
\newblock In N.~Wang, G.~Rebolledo-Mendez, N.~Matsuda, O.~C. Santos, and V.~Dimitrova, editors, {\em Artificial {Intelligence} in {Education}}, pages 627--638, Cham, 2023. Springer Nature Switzerland.

\bibitem{shi_kc-finder_2023}
Y.~Shi, R.~Schmucker, M.~Chi, T.~Barnes, and T.~Price.
\newblock {KC}-{Finder}: {Automated} {Knowledge} {Component} {Discovery} for {Programming} {Problems}.
\newblock Technical report, International Educational Data Mining Society, 2023.
\newblock ERIC Number: ED630850.

\bibitem{stamper_kc}
J.~C. Stamper and K.~R. Koedinger.
\newblock Human-machine student model discovery and improvement using datashop.
\newblock In G.~Biswas, S.~Bull, J.~Kay, and A.~Mitrovic, editors, {\em Artificial Intelligence in Education}, pages 353--360, Berlin, Heidelberg, 2011. Springer Berlin Heidelberg.

\bibitem{datashop}
J.~C. Stamper, K.~R. Koedinger, R.~S. J.~d. Baker, A.~Skogsholm, B.~Leber, S.~Demi, S.~Yu, and D.~Spencer.
\newblock Datashop: A data repository and analysis service for the learning science community (interactive event).
\newblock In G.~Biswas, S.~Bull, J.~Kay, and A.~Mitrovic, editors, {\em Artificial Intelligence in Education}, pages 628--628, Berlin, Heidelberg, 2011. Springer Berlin Heidelberg.

\bibitem{steinley_properties_2004}
D.~Steinley.
\newblock Properties of the {Hubert}-{Arable} {Adjusted} {Rand} {Index}.
\newblock {\em Psychological Methods}, 9(3):386--396, 2004.

\bibitem{doi:10.1177/1555343412474821}
C.~Tofel-Grehl and D.~F. Feldon.
\newblock Cognitive task analysis–based training: A meta-analysis of studies.
\newblock {\em Journal of Cognitive Engineering and Decision Making}, 7(3):293--304, 2013.

\bibitem{touvron2023llama2}
H.~Touvron, L.~Martin, K.~Stone, P.~Albert, A.~Almahairi, Y.~Babaei, N.~Bashlykov, S.~Batra, P.~Bhargava, S.~Bhosale, D.~Bikel, L.~Blecher, C.~C. Ferrer, M.~Chen, G.~Cucurull, D.~Esiobu, J.~Fernandes, J.~Fu, W.~Fu, B.~Fuller, C.~Gao, V.~Goswami, N.~Goyal, A.~Hartshorn, S.~Hosseini, R.~Hou, H.~Inan, M.~Kardas, V.~Kerkez, M.~Khabsa, I.~Kloumann, A.~Korenev, P.~S. Koura, M.-A. Lachaux, T.~Lavril, J.~Lee, D.~Liskovich, Y.~Lu, Y.~Mao, X.~Martinet, T.~Mihaylov, P.~Mishra, I.~Molybog, Y.~Nie, A.~Poulton, J.~Reizenstein, R.~Rungta, K.~Saladi, A.~Schelten, R.~Silva, E.~M. Smith, R.~Subramanian, X.~Tan, B.~Tang, R.~Thakoor, P.~Trinh, T.-H. Tsai, X.~Wang, W.~Wang, Z.~Wu, Y.~Zhang, M.~Zhang, P.~Zheng, M.~Zhou, and W.~Zhu.
\newblock Llama 2: Open foundation and fine-tuned chat models.
\newblock {\em arXiv preprint arXiv:2307.09288}, July 2023.

\bibitem{10.1145/1553374.1553511}
N.~X. Vinh, J.~Epps, and J.~Bailey.
\newblock Information theoretic measures for clusterings comparison: is a correction for chance necessary?
\newblock In {\em Proceedings of the 26th Annual International Conference on Machine Learning}, ICML '09, page 1073–1080, New York, NY, USA, 2009. Association for Computing Machinery.

\bibitem{wolf-etal-2020-transformers}
T.~Wolf, L.~Debut, V.~Sanh, J.~Chaumond, C.~Delangue, A.~Moi, P.~Cistac, T.~Rault, R.~Louf, M.~Funtowicz, J.~Davison, S.~Shleifer, P.~von Platen, C.~Ma, Y.~Jernite, J.~Plu, C.~Xu, T.~Le~Scao, S.~Gugger, M.~Drame, Q.~Lhoest, and A.~Rush.
\newblock Transformers: State-of-the-art natural language processing.
\newblock In Q.~Liu and D.~Schlangen, editors, {\em Proceedings of the 2020 Conference on Empirical Methods in Natural Language Processing: System Demonstrations}, pages 38--45, Online, Oct. 2020. Association for Computational Linguistics.

\bibitem{wu2016googlesneuralmachinetranslation}
Y.~Wu, M.~Schuster, Z.~Chen, Q.~V. Le, M.~Norouzi, W.~Macherey, M.~Krikun, Y.~Cao, Q.~Gao, K.~Macherey, J.~Klingner, A.~Shah, M.~Johnson, X.~Liu, Łukasz Kaiser, S.~Gouws, Y.~Kato, T.~Kudo, H.~Kazawa, K.~Stevens, G.~Kurian, N.~Patil, W.~Wang, C.~Young, J.~Smith, J.~Riesa, A.~Rudnick, O.~Vinyals, G.~Corrado, M.~Hughes, and J.~Dean.
\newblock Google's neural machine translation system: Bridging the gap between human and machine translation, 2016.

\end{thebibliography}

\balancecolumns
% That's all folks!
\end{document}